\documentclass[a4paper]{article}

\usepackage{url,hyperref,lineno,microtype,subcaption}
\usepackage[onehalfspacing]{setspace}
\usepackage{graphicx}
\usepackage{amsmath}
\usepackage{amssymb}
\usepackage{booktabs}
\usepackage{multirow}
\usepackage{graphicx}
\usepackage{floatrow}
\usepackage{caption}
\usepackage{wasysym}
\usepackage{wrapfig}
\linenumbers

\title{Heterogeneous Recurrent Spiking Neural Network for Spatio-Temporal Classification}

\author{Biswadeep Chakraborty\,$^{1,*}$ and Saibal Mukhopadhyay\,$^{1}$}

\date{$^{1}$ Georgia Institute of Technology, Department of Electrical and Computer Engineering, Atlanta, GA, USA}

\begin{document}
\nolinenumbers

\maketitle

\begin{abstract}
Spiking Neural Networks are often touted as brain-inspired learning models for the third wave of Artificial Intelligence. Although recent SNNs trained with supervised backpropagation show classification accuracy comparable to deep networks, the performance of unsupervised learning-based SNNs remains much lower. This paper presents a heterogeneous recurrent spiking neural network (HRSNN) with unsupervised learning for spatio-temporal classification of video activity recognition tasks on RGB (KTH, UCF11, UCF101) and event-based datasets (DVS128 Gesture). The key novelty of the HRSNN is that the recurrent layer in HRSNN consists of heterogeneous neurons with varying firing/relaxation dynamics, and they are trained via heterogeneous spike-time-dependent-plasticity (STDP) with varying learning dynamics for each synapse. We show that this novel combination of heterogeneity in architecture and learning method outperforms current homogeneous spiking neural networks. We further show that HRSNN can achieve similar performance to state-of-the-art backpropagation trained supervised SNN, but with less computation (fewer neurons and sparse connection) and less training data. 

\end{abstract}

\section{Introduction}

Acclaimed as the third generation of neural networks, spiking neural networks (SNNs) have become very popular. In general, SNN promises lower operating power when mapped to hardware. In addition, recent developments of SNNs with leaky integrate-and-fire (LIF) neurons have shown classification performance similar to deep neural networks (DNN). However, most of these works use supervised statistical training algorithms such as backpropagation-through-time (BPTT) \cite{wu2018spatio}, \cite{jin2018hybrid}, \cite{shrestha2018slayer}. These backpropagated models are extremely data-dependent and show poor trainability with less training data, and generalization characteristics \cite{lobo2020spiking}, \cite{tavanaei2019deep}. In addition, BPTT-trained models need highly complex architecture with a large number of neurons for good performance. Though unsupervised learning methods like the STDP have been introduced, they lack performance compared to their backpropagated counterparts. This is attributed to the high training complexity of these STDP dynamics \cite{lazar2006combination}. Therefore, there is a need to explore SNN architectures and algorithms that can improve the performance of unsupervised learned SNN. 
\par This paper introduces a Heterogeneous Recurrent Spiking Neural Network (HRSNN) with heterogeneity in both the LIF neuron parameters and the STDP dynamics between the neurons. Recent works have discussed that heterogeneity in neuron time constants improves the model's performance in the classification task \cite{perez2021neural},\cite{she2021sequence},\cite{yin2021accurate},\cite{zeldenrust2021efficient}. However, these papers lack a theoretical understanding of why heterogeneity improves the classification properties of the network and thus, also fails to explore the various advantages that arise from it, like training with less data or sparsity in neurons and network connections. Also, they do not explore the advantages of heterogeneity in both the neuronal hyperparameters and the learning dynamics, limiting the network's capabilities. Our work also uses a novel BO method to optimize the hyperparameter search process, making it highly scalable for larger heterogeneous networks that can be used for more complex tasks like action recognition, which was not possible earlier. 
First, we analytically show that heterogeneity improves the linear separation property of unsupervised SNN models. We also empirically verified that heterogeneity in the LIF parameters and the STDP dynamics significantly improves the classification performance using fewer neurons, sparse connections, and lesser training data. We use a Bayesian Optimization (BO)-based method using a modified Matern Kernel on the Wasserstein metric space to search for optimal parameters of the HRSNN model and evaluate the performance on RGB (KTH, UCF11, UCF101) and event-based datasets (DVS-Gesture). The HRSNN model achieves an accuracy of 94.32\% on KTH, 79.58\% on UCF11, 77.33\% on UCF101, and 96.54\% on DVS-Gesture using 2000 LIF neurons.

\section{Related Works}

\subsection{Recurrent Spiking Neural Network}
\par \textbf{Supervised Learning: } Recurrent networks of spiking neurons can be effectively trained to achieve competitive performance compared to standard recurrent neural networks. Demin et al. \cite{demin2018recurrent} showed that using recurrence could reduce the number of layers in SNN models and potentially form the various functional network structures. Zhang et al. \cite{zhang2019spike} proposed a spike‐train level recurrent SNN backpropagation method to train the deep RSNNs, which achieves excellent performance in image and speech classification tasks. On the other hand, Wang et al. \cite{wang2021recurrent} used the recurrent LIF neuron model with the dynamic presynaptic currents and trained by the BP based on surrogate gradient. 

\textbf{Unsupervised Learning: } Unsupervised learning models like STDP have shown great generalization and trainability properties \cite{chakraborty2021characterization}. Previous works have used STDP for training the recurrent spiking networks \cite{gilson2010stdp}. Nobukawa et al. \cite{nobukawa2019pattern} used a hybrid STDP and Dopamine-modulated STDP to train the recurrent spiking network and showed its performance in classifying patterns. Several other works have used a reservoir-based computing strategy as described above. Liquid State Machines, equipped with unsupervised learning models like STDP and BCM \cite{ivanov2021increasing} have shown promising results. 
\par \textbf{Heterogeneity: } Despite the previous works on recurrent spiking neural networks, all these models use a uniform parameter distribution for spiking neuron parameters and their learning dynamics. There has been little research leveraging heterogeneity in the model parameters and their effect on the performance and generalization. Recently, Perez-Nieves et al. \cite{perez2021neural} introduced heterogeneity in the neuron time-constants and showed this improves the model's performance in the classification task and makes the model robust to hyperparameter tuning. She et al. \cite{she2021sequence} also used a similar heterogeneity in the model parameters of a feedforward spiking network and showed it could classify temporal sequences. Again, modeling heterogeneity in the brain cortical networks, Zeldenrust et al. derived a class of RSNNs that tracks a continuously varying input online \cite{zeldenrust2021efficient}.

\subsection{Action Recognition using SNNs}
SNNs cann operate directly on the event data instead of aggregating them, recent works use the concept of time-surfaces \cite{maro2020event} \cite{lagorce2016hots}. Recent works have used convolution and reservoir computing techniques like liquid state machines to solve the event-based gesture recognition problem \cite{george2020reservoir} \cite{zhou2020surrogate} \cite{panda2018learning}.

\section{Methods}
\subsection{Recurrent Spiking Neural Network}

\begin{figure*}
    \centering
    \includegraphics[width = \textwidth]{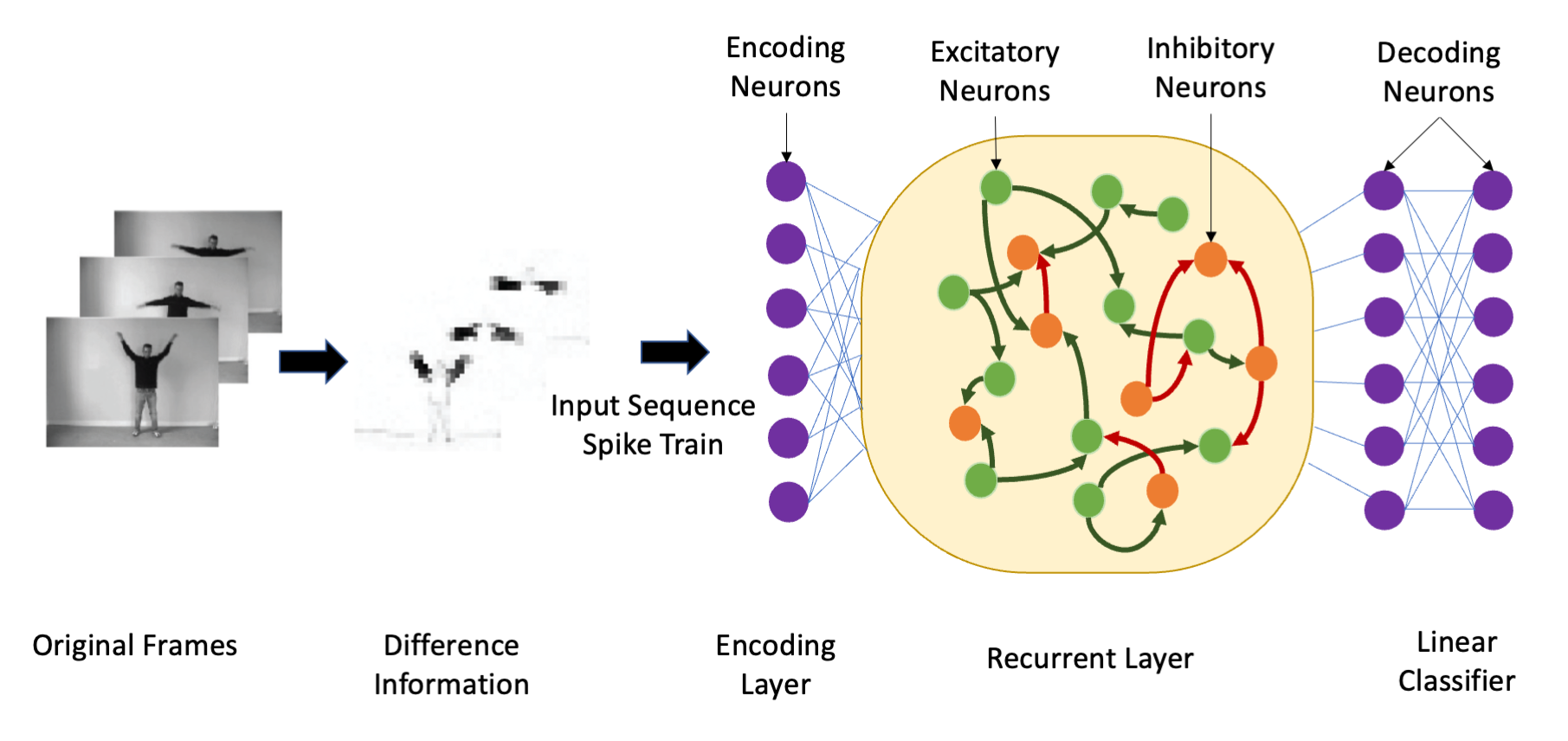}
    \caption{An illustrative example showing the Heterogeneous Recurrent Spiking Neural Network structure. First, we show the temporal encoding method based on the sensory receptors receiving the difference between two time-adjacent data. Next, the input sequences are encoded by the encoding neurons that inject the spike train into $30\%$ neurons in $\mathcal{R}. \mathcal{R}$ contains a $4:1$ ratio of excitatory (green nodes) and inhibitory (orange nodes), where the neuron parameters are heterogeneous. The synapses are trained using the heterogeneous STDP method.}
    \label{fig:model}
\end{figure*}

SNN consists of spiking neurons connected with synapses. The spiking LIF is defined by the following equations:
\begin{equation}
    \tau_{m} \frac{d v}{d t}=a+R_{m} I-v ; v=v_{\text {reset }}, \text { if } v>v_{\text {threshold }}
\end{equation}
where $R_{m}$ is membrane resistance, $\tau_{m}=R_{m} C_{m}$ is time constant and $C_{m}$ is membrane capacitance. $a$ is the resting potential. $I$ is the sum of current from all input synapses connected to the neuron. A spike is generated when membrane potential $v$ crosses the threshold, and the neuron enters refractory period $r$, during which the neuron maintains its membrane potential at $v_{\text {reset }}.$
We construct the HRSNN from the baseline recurrent spiking network (RSNN) consisting of three layers: (1) an input encoding layer ($\mathcal{I}$), (2) a recurrent spiking layer ($\mathcal{R}$), and (3) an output decoding layer ($\mathcal{O}$). The recurrent layer consists of excitatory and inhibitory neurons, distributed in a ratio of $N_{E}: N_{I}=4: 1$. The PSPs of post-synaptic neurons produced by the excitatory neurons are positive, while those produced by the inhibitory neurons are negative. We used a biologically plausible LIF neuron model and trained the model using STDP rules. 

From here on, we refer to connections between $\mathcal{I}$ and $\mathcal{R}$ neurons as $\mathcal{S}_{\mathcal{IR}}$ connections, inter-recurrent layer connections as $\mathcal{S}_{\mathcal{RR}}$, and $\mathcal{R}$ to $\mathcal{O}$ as $\mathcal{S}_{\mathcal{RO}}$. We created $\mathcal{S}_{\mathcal{RR}}$ connections using probabilities based on Euclidean distance, $D(i, j)$, between any two neurons $i, j$:
\begin{equation}
P(i, j)=C \cdot \exp \left(-\left(\frac{D(i, j)}{\lambda}\right)^{2}\right)
\label{eq:eucl}
\end{equation}
with closer neurons having higher connection probability. Parameters $C$ and $\lambda$ set the amplitude and horizontal shift, respectively, of the probability distribution.
$\mathcal{I}$ contains excitatory encoding neurons, which convert input data into spike trains. $\mathcal{S}_{IR}$ only randomly chooses $30\%$ of the excitatory and inhibitory neurons in $\mathcal{R}$ as the post-synaptic neuron. The connection probability between the encoding neurons and neurons in the $\mathcal{R}$ is defined by a uniform probability $\mathcal{P}_{\mathcal{IR}}$, which, together with $\lambda$, will be used to encode the architecture of the HRSNN and optimized using BO. In this work, each neuron received projections from some randomly selected neurons in $\mathcal{R}$. 

We used unsupervised, local learning to the spiking recurrent model by letting STDP change each $\mathcal{S}_{\mathcal{RR}}$ and $\mathcal{S}_{\mathcal{IR}}$ connection, modeled as:
\begin{equation}
\frac{d W}{d t}=A_{+} T_{p r e} \sum_{o} \delta\left(t-t_{\text {post }}^{o}\right)-A_{-} T_{\text {post }} \sum_{i} \delta\left(t-t_{\text {pre }}^{i}\right)
\end{equation}
where $A_{+}, A_{-}$ are the potentiation/depression learning rates and $T_{\text {pre }} / T_{\text {post }}$ are the pre/post-synaptic trace variables, modeled as,
\begin{align}
\tau_{+}^{*} \frac{d T_{\text {pre }}}{d t} &=-T_{\text {pre }}+a_{+} \sum_{i} \delta\left(t-t_{\text {pre }}^{i}\right) \\
\tau_{-}^{*} \frac{d T_{\text {post }}}{d t} &=-T_{\text {post }}+a_{-} \sum_{o} \delta\left(t-t_{\text {post }}^{o}\right)
\end{align}
where $a_{+}, a_{-}$ are the discrete contributions of each spike to the trace variable, $\tau_{+}^{*}, \tau_{-}^{*}$ are the decay time constants, $t_{\text {pre }}^{i}$ and $t_{\text {post }}^{o}$ are the times of the pre-synaptic and post-synaptic spikes, respectively.

\textbf{Heterogeneous LIF Neurons} \label{sec:het_lif}
The use of multiple timescales in spiking neural networks has several underlying benefits, like increasing the memory capacity of the network. In this paper, we propose the usage of heterogeneous LIF neurons with different membrane time constants and threshold voltages, thereby giving rise to multiple timescales. Due to differential effects of excitatory and inhibitory heterogeneity on the gain and asynchronous state of sparse cortical networks \cite{carvalho2009differential}, \cite{hofer2011differential}, we use different gamma distributions for both the excitatory and inhibitory LIF neurons. This is also inspired by the brain's biological observations, where the time constants for excitatory neurons are larger than the time constants for the inhibitory neurons. Thus, we incorporate the heterogeneity in our Recurrent Spiking Neural Network by using different membrane time constants $\tau$ for each LIF neuron in $\mathcal{R}$. This gives rise to a distribution for the time constants of the LIF neurons in $\mathcal{R}$.

\textbf{Heterogeneous STDP} \label{sec:het_stdp}
Experiments on different brain regions and diverse neuronal types have revealed a wide variety of STDP forms that vary in plasticity direction, temporal dependence, and the involvement of signaling pathways \cite{sjostrom2008dendritic}, \cite{feldman2012spike}, \cite{korte2016cellular}.
As described by Pool et al. \cite{pool2011spike}, one of the most striking aspects of this plasticity mechanism in synaptic efficacy is that the STDP windows display a great variety of forms in different parts of the nervous system. However, most STDP models used in Spiking Neural Networks are homogeneous with uniform timescale distribution. Thus, we explore the advantages of using heterogeneities in several hyperparameters discussed above. This paper considers heterogeneity in the scaling function constants ($A_+, A_-$) and the decay time constants ($\tau_+, \tau_-$).

\subsection{Classification Property of HRSNN}

We theoretically compare the performance of the heterogeneous spiking recurrent model with its homogeneous counterpart using a binary classification problem. The ability of HRSNN to distinguish between many inputs is studied through the lens of the edge-of-chaos dynamics of the spiking recurrent neural network, similar to the case in spiking reservoirs shown by Legenstein et al. \cite{legenstein2007edge}. Also, $\mathcal{R}$ possesses a fading memory due to its short-term synaptic plasticity and recurrent connectivity. For each stimulus, the final state of the $\mathcal{R}$, i.e., the state at the end of each stimulus, carries the most information. The authors showed that the rank of the final state matrix $F$ reflects the separation property of a kernel: $F=\left[S(1) \quad S(2) \quad \cdots \quad S(N) \right]^T$ where $S(i)$ is the final state vector of $\mathcal{R}$ for the stimulus $i$. Each element of $F$ represents one neuron's response to all the $N$ stimuli. A higher rank in $F$ indicates better kernel separation if all $N$ inputs are from $N$ distinct classes.

The effective rank is calculated using Singular Value Decomposition (SVD) on $F$, and then taking the number of singular values that contain $99 \%$ of the sum in the diagonal matrix as the rank. i.e. $F=U \Sigma V^{T}$
where $U$ and $V$ are unitary matrices, and $\Sigma$ is a diagonal matrix $\operatorname{diag}\left(\lambda_{1}, \lambda_{2}, \lambda_{3}, \ldots, \lambda_{N}\right)$ that contains non-negative singular values such that $(\lambda_1 \ge \lambda_2 \cdots \ge \lambda_N)$.

\textbf{Definition: \textit{Linear separation property}} \textit{of a neuronal circuit $\mathcal{C}$ for $m$ different inputs $u_{1}, \ldots, u_{m}(t)$  is defined as the rank of the $n \times m$ matrix $M$ whose columns are the final circuit states $\mathbf{x}_{u_{i}}\left(t_{0}\right)$ obtained at time $t_{0}$ for the preceding input stream $u_{i}$.}

Following from the definition introduced by Legenstein et al. \cite{legenstein2007edge}, if the rank of the matrix $M= m$, then for the inputs $u_{i}$, any given assignment of target outputs $y_{i} \in \mathbb{R}$ at time $t_{0}$  can be implemented by $\mathcal{C}$.

We use the rank of the matrix as a measure for the linear separation of a circuit $C$  for distinct inputs. This leverages the complexity and diversity of nonlinear operations carried out by $C$ on its input to boost the classification performance of a subsequent linear decision-hyperplane.

\textit{\textbf{Theorem 1: }  Assuming $\mathcal{S}_u$ is finite and contains $s$ inputs, let $r_{\text{Hom}}, r_{\text{Het}}$ are the ranks of the $n \times s$ matrices consisting of the $s$ vectors $\mathbf{x}_{u}\left(t_{0}\right)$ for all inputs $u$ in $\mathcal{S}_u$ for each of Homogeneous and Heterogeneous RSNNs respectively. Then $r_{\text{Hom}} \leq r_{\text{Het}}$.}

\textit{\textbf{Short Proof:}} Let us fix some inputs $u_{1}, \ldots, u_{r}$ in $\mathcal{S}_u$ so that the resulting $r$ circuit states $\mathbf{x}_{u_{i}}\left(t_{0}\right)$ are linearly independent. Using the Eckart-Young-Mirsky theorem for low-rank approximation, we show that the number of linearly independent vectors for HeNHeS is greater than or equal to the number of linearly independent vectors for HoNHoS. The detailed proof is given in the Supplementary.

\textbf{Definition :}  \textit{Given $K_{\rho}$ is the modified Bessel function of the second kind, and $\sigma^{2}, \kappa, \rho$ are the variance, length scale, and smoothness parameters respectively, we define the \textbf{modified Matern kernel on the Wasserstein metric space} $\mathcal{W}$ between two distributions $\mathcal{X}, \mathcal{X}^{\prime}$ given as}
\begin{equation}
    k\left(\mathcal{X}, \mathcal{X}^{\prime}\right)=\sigma^{2} \frac{2^{1-\rho}}{\Gamma(\rho)}\left(\sqrt{2 \rho} \frac{\mathcal{W}(\mathcal{X}, \mathcal{X}^{\prime})}{\kappa}\right)^{\rho} H_{\rho}\left(\sqrt{2 \rho} \frac{\mathcal{}(\mathcal{X}, \mathcal{X}^{\prime})}{\kappa}\right)
\label{eq:matern}
\end{equation}

where $\Gamma(.), H(.)$ is the Gamma and Bessel function, respectively.

\textit{\textbf{Theorem 2:} The modified Matern function on the Wasserstein metric space $\mathcal{W}$ is a valid kernel function}

\textbf{Short Proof: } To show that the above function is a kernel function, we need to prove that Mercer's theorem holds. i.e., (i) the function is symmetric and (ii) in finite input space, the Gram matrix of the kernel function is positive semi-definite. The detailed proof is given in the Supplementary.

\subsection{Optimal Hyperparameter Selection using Bayesian Optimization}
While BO is used in various settings, successful applications are often limited to low-dimensional problems, with fewer than twenty dimensions \cite{frazier2018tutorial}. Thus, using BO for high-dimensional problems remains a significant challenge. In our case of optimizing HRSNN model parameters for 2000, we need to optimize a huge number of parameters, which is extremely difficult for BO. As discussed by Eriksson et al. \cite{eriksson2021high}, suitable function priors are especially important for good performance. Thus, we used a biologically inspired initialization of the hyperparameters derived from the human brain (see Supplementary for details). 

This paper uses a  modified BO to estimate parameter distributions for the LIF neurons and the STDP dynamics. To learn the probability distribution of the data, we modify the surrogate model and the acquisition function of the BO to treat the parameter distributions instead of individual variables. This makes our modified BO highly scalable over all the variables (dimensions) used. The loss for the surrogate model's update is calculated using the Wasserstein distance between the parameter distributions.

BO uses a Gaussian process to model the distribution of an objective function and an acquisition function to decide points to evaluate. For data points in a target dataset $x \in X$ and the corresponding label $y \in Y$, an SNN with network structure $\mathcal{V}$ and neuron parameters $\mathcal{W}$ acts as a function $f_{\mathcal{V}, \mathcal{W}}(x)$ that maps input data $x$ to predicted label $\tilde{y}$. The optimization problem in this work is defined as
\begin{equation}
    \min _{\mathcal{V}, \mathcal{W}} \sum_{x \in X, y \in Y} \mathcal{L}\left(y, f_{\mathcal{V}, \mathcal{W}}(x)\right)
\end{equation}
where $\mathcal{V}$ is the set of hyperparameters of the neurons in $\mathcal{R}$ (Details of hyperparameters given in the Supplementary) and $\mathcal{W}$ is the multi-variate distribution constituting the distributions of: (i) the membrane time constants $\tau_{m-E}, \tau_{m-I}$ of the LIF neurons, (ii) the scaling function constants $(A_+, A_-)$ and (iii) the decay time constants $\tau_+, \tau_-$ for the STDP learning rule in $\mathcal{S}_{\mathcal{RR}}$.

Again, BO needs a prior distribution of the objective function $f(\vec{x})$ on the given data $\mathcal{D}_{1: k}=\left\{\vec{x}_{1: k}, f\left(\vec{x}_{1: k}\right)\right\}.$ 
In GP-based BO, it is assumed that the prior distribution of $f\left(\vec{x}_{1: k}\right)$ follows the multivariate Gaussian distribution which follows a Gaussian Process with mean $\vec{\mu}_{\mathcal{D}_{1: k}}$ and covariance $\vec{\Sigma}_{\mathcal{D}_{1: k}}$. We estimate $\vec{\Sigma}_{\mathcal{D}_{1: k}}$ using the modified Matern kernel function which is given in Eq. \ref{eq:matern}. In this paper, we use $d(x,x^{\prime})$ as the Wasserstein distance between the multivariate distributions of the different parameters. It is to be noted here that for higher-dimensional metric spaces, we use the Sinkhorn distance as a regularized version of the Wasserstein distance to approximate the Wasserstein distance \cite{feydy2019interpolating}.  

$\mathcal{D}_{1: k}$ are the points that have been evaluated by the objective function and the GP will estimate the mean $\vec{\mu}_{\mathcal{D}_{k: n}}$ and variance $\vec{\sigma}_{\mathcal{D}_{k: n}}$ for the rest unevaluated data $\mathcal{D}_{k: n}$. The acquisition function used in this work is the expected improvement (EI) of the prediction fitness, as:
\begin{equation}
E I\left(\vec{x}_{k: n}\right)=\left(\vec{\mu}_{\mathcal{D}_{k: n}}-f\left(x_{\text {best }}\right)\right) \Phi(\vec{Z})+\vec{\sigma}_{\mathcal{D}_{k: n}} \phi(\vec{Z})
\end{equation}
where $\Phi(\cdot)$ and $\phi(\cdot)$ denote the probability distribution function and the cumulative distribution function of the prior distributions, respectively. $f\left(x_{\text {best }}\right)=\max f\left(\vec{x}_{1: k}\right)$ is the maximum value that has been evaluated by the original function $f$ in all evaluated data $\mathcal{D}_{1: k}$ and $\vec{Z}=\frac{\vec{\mu}_{\mathcal{D}_{k: n}}-f\left(x_{\text {best }}\right)}{\vec{\sigma}_{\mathcal{D}_{k: n}}}$. BO will choose the data $x_{j}=$ $\operatorname{argmax}\left\{E I\left(\vec{x}_{k: n}\right) ; x_{j} \subseteq \vec{x}_{k: n}\right\}$ as the next point to be evaluated using the original objective function. 

\section{Experiments}

\begin{figure}[b]
    \centering
    \includegraphics[width=0.9\columnwidth]{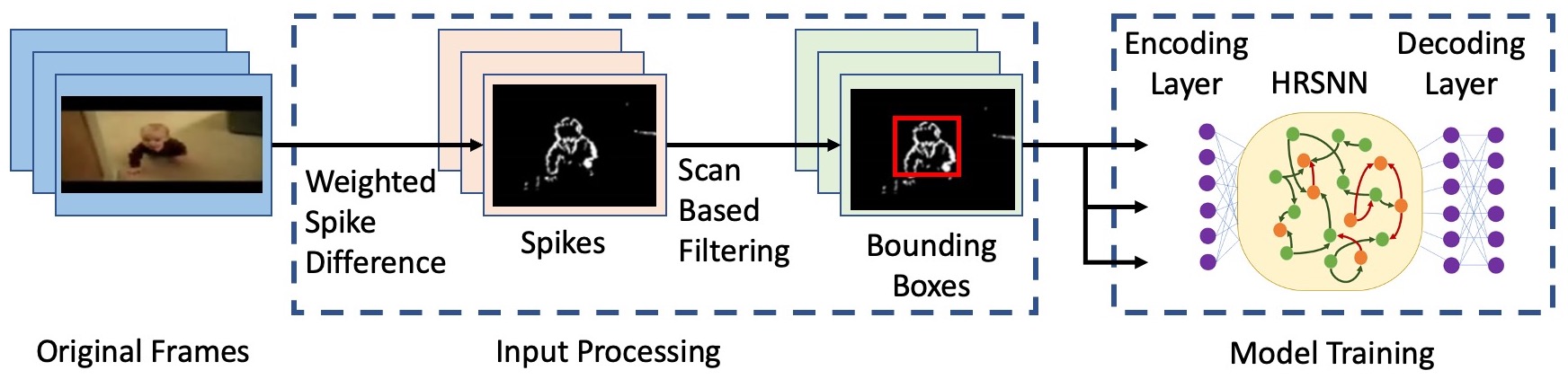}
    \caption{Figure showing a flowchart for the input processing and model training}
    \label{fig:4}
\end{figure}

\subsection{Training and Inference}
First, to remove the background noise which arises due to camera movement and jitters, we use the Scan-based filtering technique as proposed by Panda et al. \cite{panda2018learning} where we create a bounding box and center of gravity of spiking activity for each frame and scan across five directions as shown in Fig. \ref{fig:4}.
The recurrent spiking layer extracts the features of the spatio-temporal data and converts them into linearly separable states in a high-dimensional space. $\mathcal{O}$ abstracts the state from $\mathcal{R}$ for classification. The state of $\mathcal{R}$ is defined as the membrane potential of the output neurons at the end of each spike train converted from the injected spatio-temporal data. After the state is extracted, the membrane potential of the output neuron is set to its initial value. After injecting all sequences into the network, the states of each data are obtained. A linear classifier is employed in this work to evaluate pattern recognition performance.

\subsection{Baseline Ablation Models}
We use the following baselines for the comparative study:
\begin{itemize}
    
        \item \textbf{Recurrent Spiking Neural Network with STDP: }
        \begin{itemize}
    \item Homogeneous LIF Neurons and Homogeneous STDP Learning (\textbf{HoNHoS})  
    
    \item Heterogeneity in LIF Neuron Parameters and Homogeneous STDP Learning (\textbf{HeNHoS}) 
    \item Homogeneous LIF Neuron Parameters and Heterogeneity in LTP/LTD dynamics of STDP  (\textbf{HoNHeS}) 
    \item Heterogeneity in both LIF and STDP parameters (\textbf{HeNHeS}) 
\end{itemize}
\item \textbf{Recurrent Spiking Neural Network with Backpropagation: }
  
\begin{itemize}
    \item Homogeneous LIF Neurons trained with Backpropagation (\textbf{HoNB})  
    \item Heterogeneous LIF Neurons trained with Backpropagation (\textbf{HeNB}) 
\end{itemize}

  \end{itemize}

 \subsection{Computational Cost:}
To estimate the efficiency of BP-SNNs and compare them with DNNs, we calculate the number of computations required in terms of accumulation (AC) and multiply-accumulation (MAC) operations \cite{wong2020tinyspeech}. In DNNs, the contribution from one neuron to another requires a MAC for every timestep where each input activation is multiplied with the respective weight before it is added to the internal sum. On the other hand, for a spiking neuron, a transmitted spike requires only an accumulation at the target neuron, adding weight to the potential, and where spikes may be quite sparse. As it is much more energetically expensive to calculate MACs than ACs (on a 45nm 0.9V chip, a 32-bit floating-point (FL) MAC operation consumes 4.6 pJ and 0.9 pJ for an AC operation \cite{chakraborty2021fully}, \cite{panda2018learning}), the relative efficiency of SNNs is determined by the number of connections multiplied by activity sparsity and the spiking neuron model complexity.

\section{Results}


\subsection{Ablation Studies}

We compare the performance of the HRSNN model with heterogeneity in the LIF and STDP dynamics (HeNHeS) to the ablation baseline recurrent spiking neural network models described above. We run five iterations for all the baseline cases and show the mean and standard deviation of the prediction accuracy of the network using 2000 neurons. The results are shown in Table \ref{tab:ablation}. We see that the heterogeneity in the LIF neurons and the LTP/LTD dynamics significantly improve the model's accuracy and error.

\begin{table*}[t]
\centering
\caption{Table comparing the performance of RSNN with Homogeneous and Heterogeneous LIF neurons using different learning methods with 2000 neurons}
\label{tab:ablation}
\resizebox{\textwidth}{!}{%
\begin{tabular}{|c|ccc|ccc|}
\hline
Datasets & \multicolumn{3}{c|}{KTH} & \multicolumn{3}{c|}{DVS128} \\ \hline
\begin{tabular}[c]{@{}c@{}}Neuron \\ Type\end{tabular} & \multicolumn{1}{c|}{\begin{tabular}[c]{@{}c@{}}Homogeneous\\ STDP\end{tabular}} & \multicolumn{1}{c|}{\begin{tabular}[c]{@{}c@{}}Heterogeneous\\ STDP\end{tabular}} & BackPropagation & \multicolumn{1}{c|}{\begin{tabular}[c]{@{}c@{}}Homogeneous\\ STDP\end{tabular}} & \multicolumn{1}{c|}{\begin{tabular}[c]{@{}c@{}}Heterogeneous\\ STDP\end{tabular}} & BackPropagation \\ \hline
\begin{tabular}[c]{@{}c@{}}Homogeneous\\ LIF\end{tabular} & \multicolumn{1}{c|}{$86.33 \pm 4.05$} & \multicolumn{1}{c|}{$91.37 \pm 3.15$} & \multicolumn{1}{c|}{$94.87 \pm 2.03$} & \multicolumn{1}{c|}{ 90.33 $\pm$ 3.41 } & \multicolumn{1}{c|}{ 93.37 $\pm$ 3.05} & 
\multicolumn{1}{c|}{ 97.06 $\pm$ 2.29}\\ \hline
\begin{tabular}[c]{@{}c@{}}Heterogeneous\\ LIF\end{tabular} & \multicolumn{1}{c|}{92.16 $\pm$ 3.17} & \multicolumn{1}{c|}{94.32 $\pm$ 1.71} & \multicolumn{1}{c|}{96.84 $\pm$ 1.96} & \multicolumn{1}{c|}{ 92.16 $\pm$ 2.97} & \multicolumn{1}{c|}{ 96.54 $\pm$ 1.82} & 
\multicolumn{1}{c|}{ 98.12 $\pm$ 1.97}\\ \hline
\end{tabular}%
}
\end{table*}

\begin{figure}[h]
    \centering
    \includegraphics[width=\textwidth]{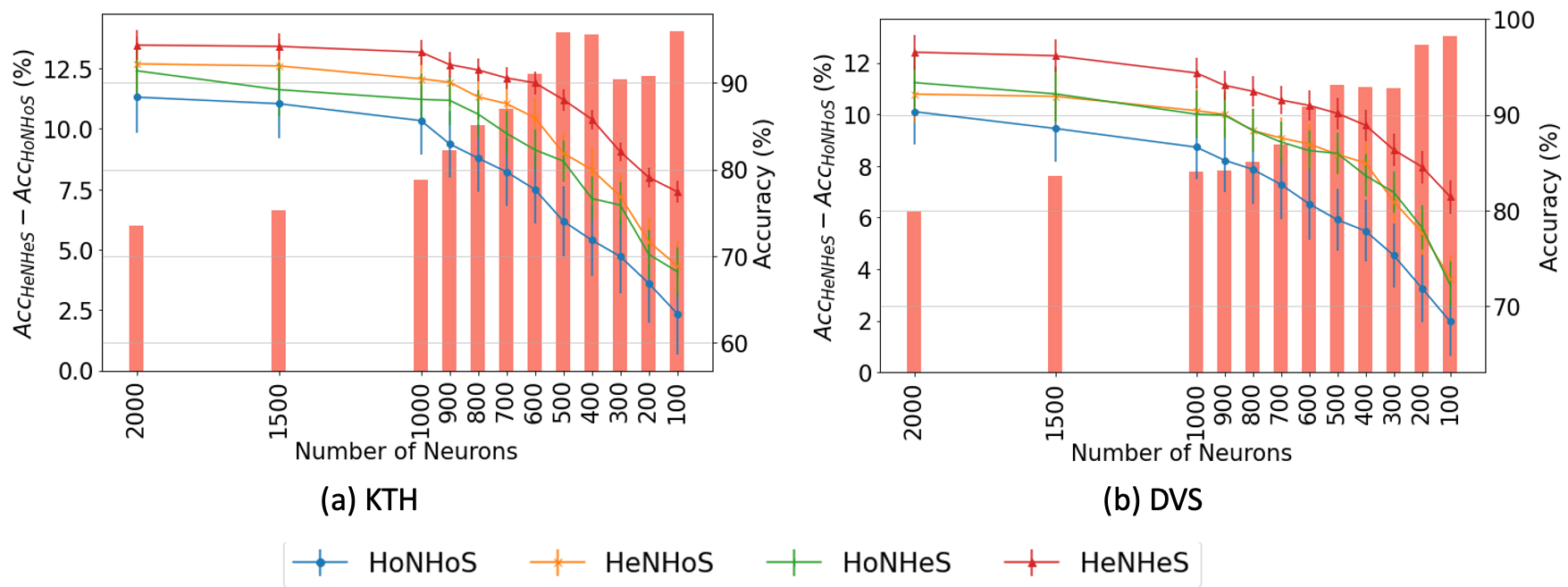}
    \caption{Comparison of performance of HRSNN models for the (a) KTH dataset and (b) DVS128 dataset for varying number of neurons. The bar graph (left Y-axis) shows the difference between the accuracies between HeNHeS and HoNHoS models. The line graphs (right Y-axis) shows the Accuracies (\%) for the four ablation networks (HoNHoS, HeNHoS, HoNHeS, HeNHeS)}
    \label{fig:acc_vs_nn}
\end{figure}

\subsection{Number of neurons}
In deep learning, it is an important task to design models with a lesser number of neurons without undergoing degradation in performance. We empirically show that heterogeneity plays a critical role in designing spiking neuron models of smaller sizes. We compare models' performance and convergence rates with fewer neurons in $\mathcal{R}$.

\par \textbf{Performance Analysis: } We analyze the network performance and error when the number of neurons is decreased from 2000 to just 100. We report the results obtained using the HoNHoS and HeNHeS models for the KTH and DVS-Gesture datasets. The experiments are repeated five times, and the observed mean and standard deviation of the accuracies are shown in Figs. \ref{fig:acc_vs_nn}. The graphs show that as the number of neurons decreases, the difference in accuracy scores between the homogeneous and the heterogeneous networks increases rapidly.

\par
\begin{figure}[h]
    \centering
    \includegraphics[width=\textwidth]{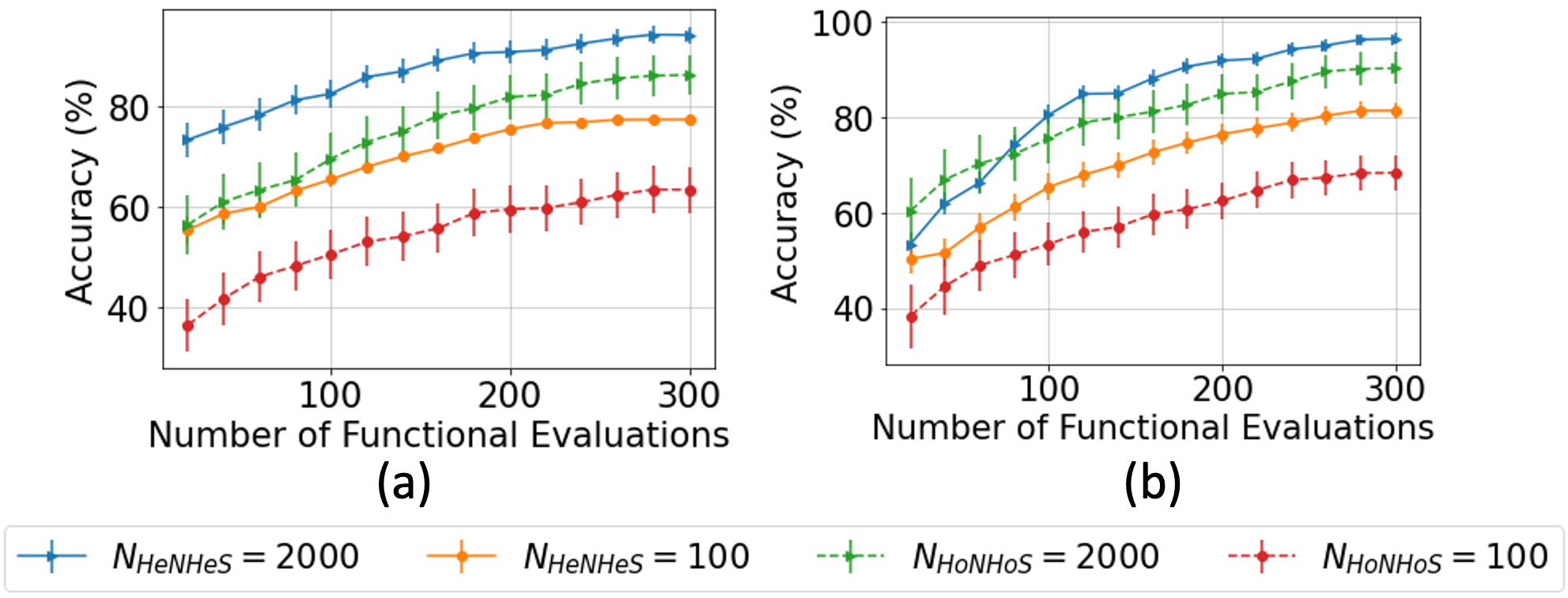}
    \caption{Plots showing comparison of the convergence of the BO with increasing functional evaluations for the (a) KTH and (b)DVSGesture dataset for varying number of neurons}
    \label{fig:acc_vs_epoch_nn}
\end{figure}

 \textbf{Convergence Analysis with lesser neurons:} Since the complexity of BO increases exponentially on increasing the search space, optimizing the HRSNN becomes increasingly difficult as the number of neurons increases. Thus, we compare the convergence behavior of the HoNHoS and HeNHeS models with 100 and 2000 neurons each.
The results are plotted in Fig. \ref{fig:acc_vs_epoch_nn}(a), (b). Despite the huge number of additional parameters, the convergence behavior of HeNHeS is similar to that of HoNHoS. Also, it must be noted that once converged, the standard deviation of the accuracies for HeNHeS is much lesser than that of HoNHoS, indicating a much more stable model.

\subsection{Sparse Connections}

\begin{figure}
    \centering
    \includegraphics[width=\linewidth]{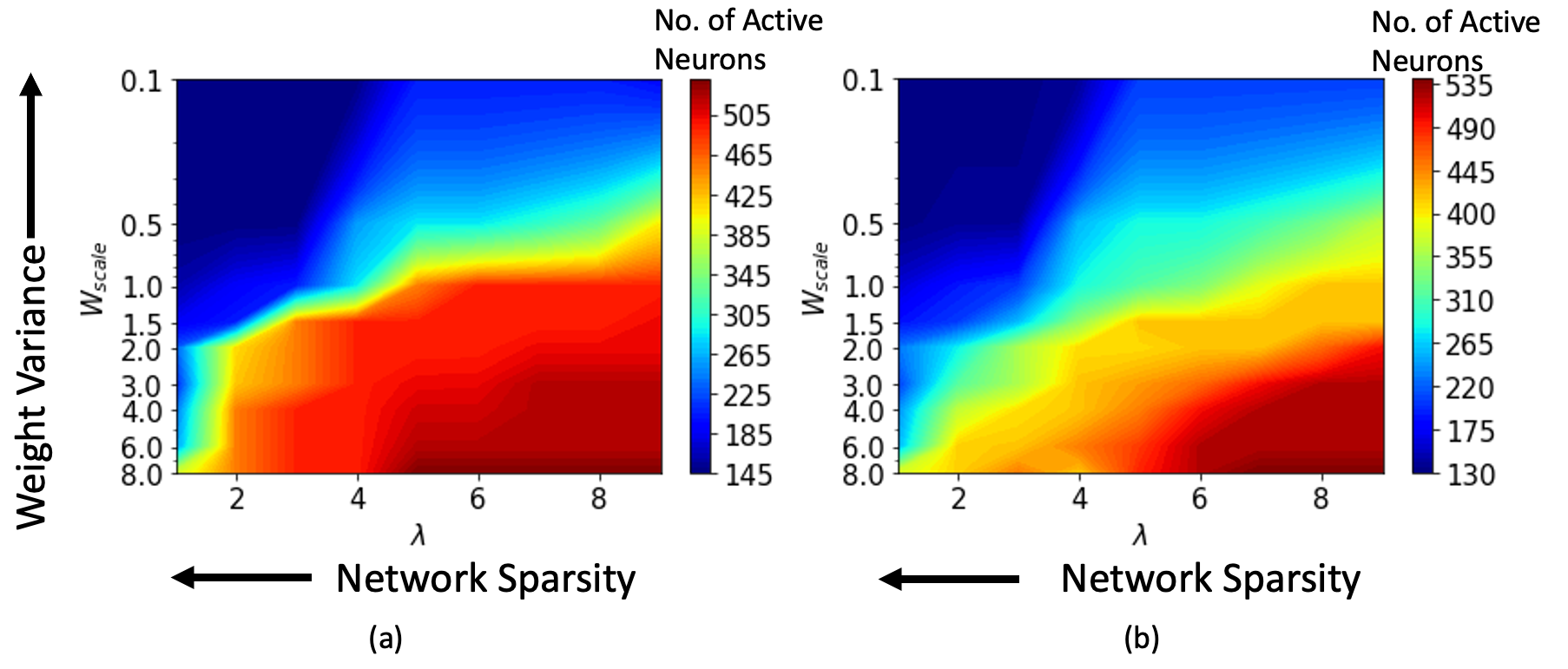}
    \caption{Change in the number of active neurons with network sparsity and weight variance, for (a)HoNHoS and (b)HeNHeS. The plot is obtained by interpolating 81 points, and each point is calculated by averaging the results from 5 randomly initialized HRSNNs.}
    \label{fig:1}
\end{figure}

$\mathcal{S}_{\mathcal{RR}}$ is generated using a probability dependent on the Euclidean distance between the two neurons, as described by Eq. \ref{eq:eucl}, where $\lambda$ controls the density of the connection, and $C$ is a constant depending on the type of the synapses \cite{zhou2020surrogate}. 

We performed various simulations using a range of values for the connection parameter $\lambda$ and synaptic weight scale $W_{\text{scale}}$. Increasing $\lambda$  will increase the number of synapses. Second, the $W_{\text{scale}}$ parameter determines the mean synaptic strength. Now, a greater $W_{\text{scale}}$ produces larger weight variance. For a single input video, the number of active neurons was calculated and plotted against the parameter values for synaptic weight $W_{\text{scale}}$ and network connectivity $\lambda$. Active neurons are those that fire at least one spike over the entire test data set. The results for the HoNHoS and HeNHeS are shown in Figs. \ref{fig:1}a, \ref{fig:1}b respectively. Each plot in the figure is obtained by interpolating 81 points, and each point is calculated by averaging the results from five randomly initialized $\mathcal {} $ with the parameters specified by the point. The horizontal axis showing the increase in $\lambda$ is plotted on a linear scale, while the vertical axis showing the variation in $W_{\text{scale}}$ is in a log scale.
The figure shows the neurons that have responded to the inputs and reflect the network's activity level. $W_{\text{scale}}$ is a factor that enhances the signal transmission within $\mathcal{R}$. As discussed by Markram et al. \cite{markram1997regulation}, the synaptic response that is generated by any action potential (AP) in a train is given as $EPSP_n = W_{\text{scale}} \times \rho_n \times u_n$, where $\rho_n$ is the fraction of the synaptic efficacy for the $n$-th AP and $u_n$ is its utilization of synaptic efficacy. Hence, it is expected that when the $W_{\text{scale}}$ is large, more neurons will fire. As $\lambda$ increases, more synaptic connections are created, which opens up more communication channels between the different neurons. As the number of synapses increases, the rank of the final state matrix used to calculate separation property also increases. The rank reaches an optimum for intermediate synapse density, and the number of synapses created increases steadily as $\lambda$ increases. As $\lambda$ increases, a larger number of connections creates more dependencies between neurons and decreases the effective separation ranks when the number of connections becomes too large. The results for the variation of the effective ranks with $\lambda$ and $W_{\text{scale}}$ are shown in the Supplementary.
\par
\begin{figure}[t]
    \centering
    \includegraphics[width=\linewidth]{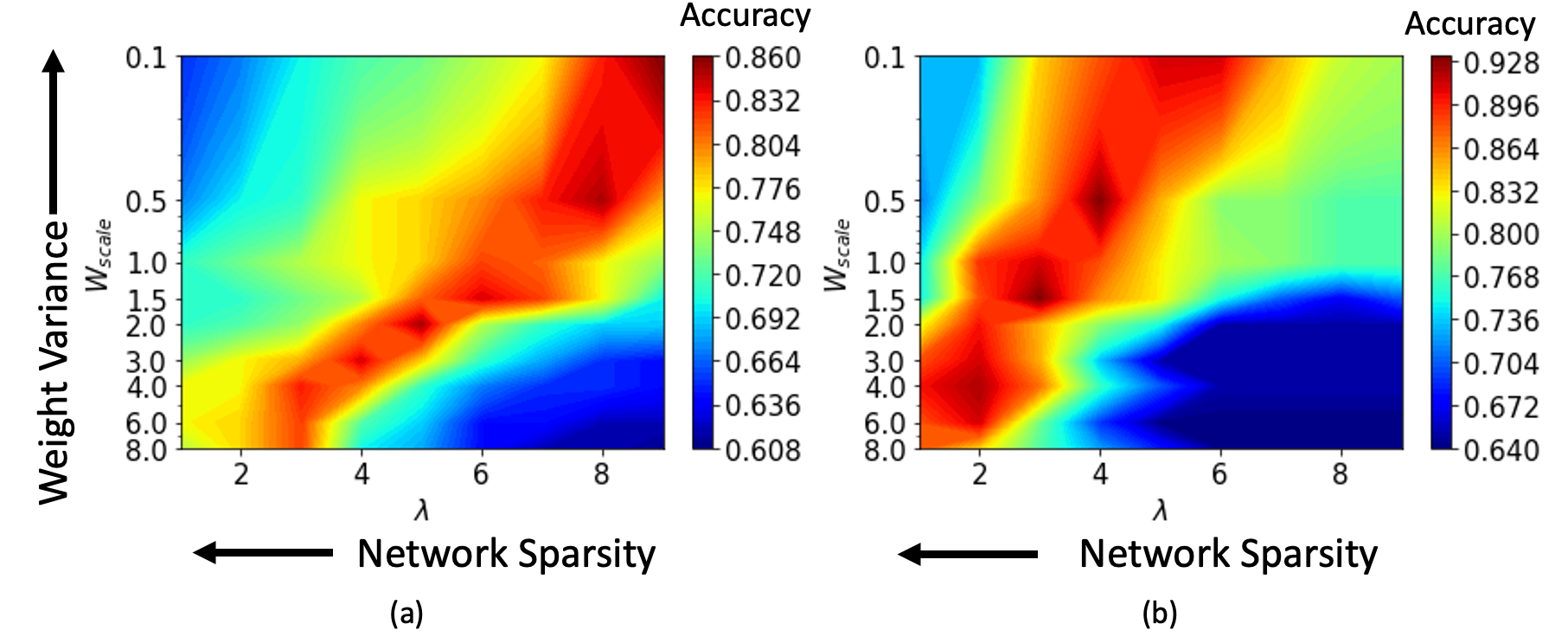}
    \caption{The variation in performance of the action recognition classification task with network sparsity and weight variance for (a)HoNHoS  and  (b)HeNHeS. The plot is obtained by interpolating 81 points, and each point is calculated by averaging the results from 5 randomly initialized HRSNNs. }
    \label{fig:3}
\end{figure}

 We compare the model's change in performance with varying sparsity levels in the connections and plotted in Figs. \ref{fig:3}a,b for the HoNHoS and the HeNHeS models. From the figures, we see that for larger values of $\lambda$, the performance of both the RSNNs was suboptimal and could not be improved by tuning the parameter $W_{\text{scale}}$. For a small number of synapses, a larger $W_{\text{scale}}$ was required to obtain satisfactory performance for HoNHoS compared to the HeNHeS model. Hence, the large variance of the weights leads to better performance. Hence, we see that the best testing accuracy for HeNHeS is achieved with fewer synapses than HoNHoS. It also explains why the highest testing accuracy for the heterogeneous network (Fig. \ref{fig:3}b.) is better than the homogeneous network (Fig. \ref{fig:3}a), because the red region in Fig. \ref{fig:3}b corresponds to higher $W_{\text{scale}}$ values and thus larger weight variance than Fig. \ref{fig:3}a.

\subsection{Limited Training Data}

\begin{figure}[t]
    \centering
    \includegraphics[width=\columnwidth]{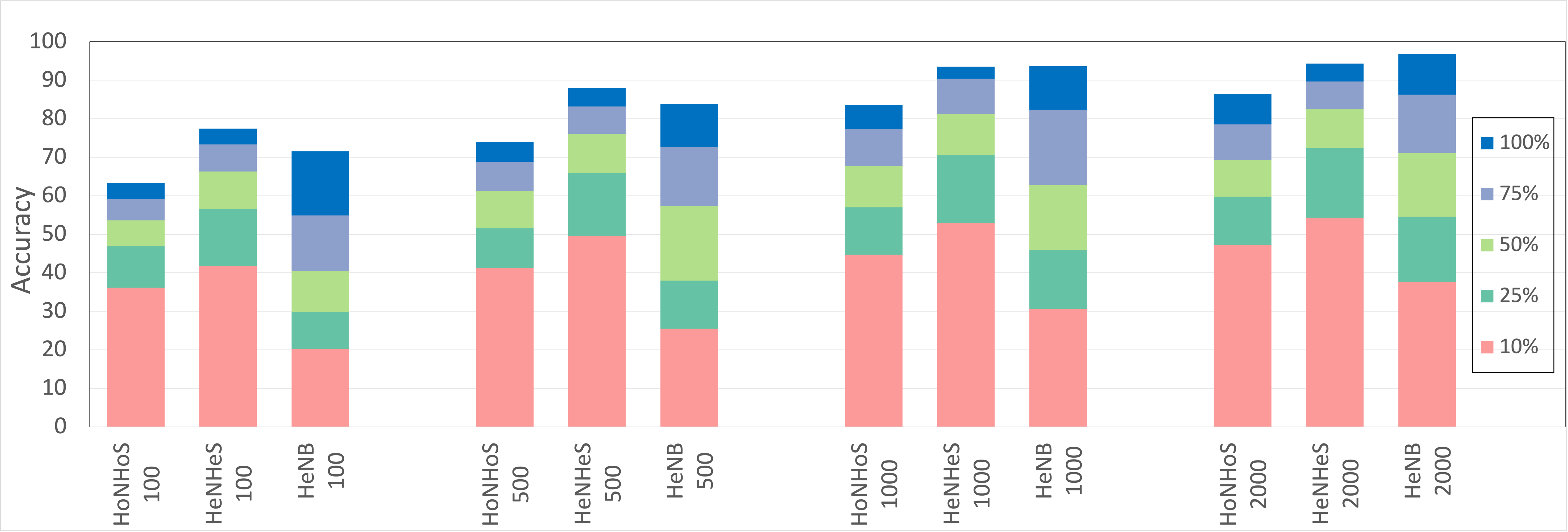}
    \caption{ Bar graph showing difference in performance for the different models with increasing training data for the KTH dataset. A similar trend can be observed for the DVS dataset (shown in Supplementary).}
    \label{fig:kth_limited}
\end{figure}

In this section, we compare the performance of the HeNHeS to HoNHoS and HeNB-based spiking recurrent neural networks that are trained with limited training data. The evaluations performed on the KTH dataset are shown in Fig. \ref{fig:kth_limited} as a stacked bar graph for the differential increments of training data sizes. The figure shows that using $10\%$ training data, HeNHeS models outperform both HoNHoS and HeNB for all the cases. The difference between the HeNHeS and HeNB increases as the number of neurons in the recurrent layer $N_\mathcal{R}$ decreases. Also, we see that adding heterogeneity improves the model's performance in homogeneous cases. Even when using 2000 neurons, HeNHeS trained with $10\%$ training data exhibit similar performance to HeNB trained with $25\%$ of training data. It is to be noted here that for the performance evaluations of the cases with $10\%$ training data, the same training was repeated until each model converged. 

\subsection{Comparison with Prior Work}

In this section, we compare our proposed HRSNN model with other baseline architectures. We divide this comparison in two parts as discussed below:

\begin{itemize}
    \item \textbf{Supervised Learning Models:} 
    \begin{itemize}
    \item We compare the performance and the model complexities of current state-of-the-art backpropagation-based SNNs with DNN-based models and their computational efficiencies. We also compare the performance of these backpropagation-based SNN models with HoNB and HeNB-based RSNN models. 
        \item \textbf{Observations: }We observe that backpropagated HRSNN models (HeNB) can achieve similar performance to DNN models but with much lesser model complexity (measured using the number of MAC/AC operations). 
    \end{itemize}
    \item \textbf{Unsupervised Learning Models:} 
    \begin{itemize}
        \item We evaluate the performance and the average spike activations for different state-of-the-art unsupervised learning models using spiking neural networks. We compare their performance with the proposed HoNHoS, HeNHeS models.
        \item \textbf{Observations: }We observe that our HRSNN models outperforms other unsupervised SNN models while using much lesser neuronal activations
    \end{itemize}
\end{itemize}

We also compare the average neuronal activation of the homogeneous and the heterogeneous recurrent SNNs for the same given image input for recurrent spiking network with 2000 neurons. If we consider the neuronal activation of neuron $i$ at time $t$ to be $\nu_i(t)$, the average neuronal activation $\bar{\nu}$ for $T$ timesteps is defined as 
 \begin{equation}
     \bar{\nu} = \frac{\Sigma_{i=0}^{N_{\mathcal{R}-1}} \Sigma_{t=0}^{T}{\nu_i(t)}}{N_\mathcal{R}} 
 \end{equation}
The results obtained are shown in Table \ref{tab:sup_comp}. The Table shows that the heterogeneous HRSNN model has a much lesser average neuronal activation than the homogeneous RSNN and the other unsupervised SNN models. Thus, we conclude that HeNHeS induces sparse activation and sparse coding of information.
Again, comparing state-of-the-art unsupervised learning models for action recognition with our proposed HRSNN models, we see that using heterogeneity in the unsupervised learning models can substantially improve the model's performance while having much lesser model complexity.

\begin{table}[]
\centering
\caption{Table showing the comparison for the Performance and the Model Complexities for DNN and Supervised and Unsupervised SNN models}
\label{tab:sup_comp}
\resizebox{\columnwidth}{!}{%
\begin{tabular}{|cccccccc|}
\hline
\multicolumn{1}{|c|}{\multirow{2}{*}{}} & \multicolumn{2}{c|}{\multirow{2}{*}{\textbf{Model}}} & \multicolumn{1}{c|}{\multirow{2}{*}{\textbf{MACs/ACs}}} & \multicolumn{3}{c|}{\textbf{RGB Dataset}} & \textbf{Event Dataset} \\ \cline{5-8} 
\multicolumn{1}{|c|}{} & \multicolumn{2}{c|}{} & \multicolumn{1}{c|}{} & \multicolumn{1}{c|}{\textbf{KTH}} & \multicolumn{1}{c|}{\textbf{UCF11}} & \multicolumn{1}{c|}{\textbf{UCF101}} & \textbf{DVS128} \\ \hline
\multicolumn{8}{|c|}{\textit{\textbf{Supervised Learning Method}}} \\ \hline
\multicolumn{1}{|c|}{\multirow{6}{*}{\textbf{DNN}}} & \multicolumn{2}{c|}{PointNet \cite{wang2019space}} & \multicolumn{1}{c|}{\textbf{MAC:} $152 \times 10^9$} & \multicolumn{1}{c|}{-} & \multicolumn{1}{c|}{-}& \multicolumn{1}{c|}{-} & 95.3 \\ \cline{2-8} 
\multicolumn{1}{|c|}{} & \multicolumn{2}{c|}{RG-CNN \cite{bi2020graph}} & \multicolumn{1}{c|}{\textbf{MAC} : $53 \times 10^9$} & \multicolumn{1}{c|}{-} & \multicolumn{1}{c|}{-} & \multicolumn{1}{c|}{-} & 97.2 \\ \cline{2-8} 
\multicolumn{1}{|c|}{} & \multicolumn{2}{c|}{I3D \cite{carreira2017quo}} & \multicolumn{1}{c|}{\textbf{MAC} : $188 \times 10^9$} & \multicolumn{1}{c|}{-} & \multicolumn{1}{c|}{90.9} & \multicolumn{1}{c|}{-} & 96.5 \\ \cline{2-8}
\multicolumn{1}{|c|}{} & \multicolumn{2}{c|}{3D-ResNet-34 \cite{lee2021low}} & \multicolumn{1}{c|}{\textbf{MAC:} $78.43 \times 10^9$} & \multicolumn{1}{c|}{94.78} & \multicolumn{1}{c|}{83.72} & \multicolumn{1}{c|}{-} & - \\ \cline{2-8} 
\multicolumn{1}{|c|}{} & \multicolumn{2}{c|}{3D-ResNet-50 \cite{lee2021low}} & \multicolumn{1}{c|}{\textbf{MAC:} $62.09 \times 10^9$} & \multicolumn{1}{c|}{92.31} & \multicolumn{1}{c|}{81.44} & \multicolumn{1}{c|}{-} & - \\ \cline{2-8} 
\multicolumn{1}{|c|}{} & \multicolumn{2}{c|}{TDN \cite{wang2021tdn}} & \multicolumn{1}{c|}{\textbf{MAC:} $69.67 \times 10^9$} & \multicolumn{1}{c|}{99.15} & \multicolumn{1}{c|}{98.03} & \multicolumn{1}{c|}{97.4} & - \\ \hline

\multicolumn{1}{|c|}{\multirow{3}{*}{\textbf{\begin{tabular}[c]{@{}c@{}}SNN-\\ Supervised\end{tabular}}}} & \multicolumn{2}{c|}{STBP-tdBN \cite{zheng2020going}} & \multicolumn{1}{c|}{\textbf{AC:} $15.13 \times 10^7$} & \multicolumn{1}{c|}{-} & \multicolumn{1}{c|}{-} & \multicolumn{1}{c|}{-} & 96.87 \\ \cline{2-8} 
\multicolumn{1}{|c|}{} & \multicolumn{2}{c|}{Shen et al. \cite{shen2021backpropagation}} & \multicolumn{1}{c|}{\textbf{AC:} $12.14 \times 10^7$} & \multicolumn{1}{c|}{-} & \multicolumn{1}{c|}{-} & \multicolumn{1}{c|}{-} & 98.26 \\ \cline{2-8} 
\multicolumn{1}{|c|}{} & \multicolumn{2}{c|}{Liu et al. \cite{liu2021event}} & \multicolumn{1}{c|}{\textbf{AC:} $27.59 \times 10^7$} & \multicolumn{1}{c|}{90.16} & \multicolumn{1}{c|}{-} & \multicolumn{1}{c|}{-} & 92.7 \\ \cline{2-8}
\multicolumn{1}{|c|}{} & \multicolumn{2}{c|}{Panda et al. \cite{panda2018learning}} & \multicolumn{1}{c|}{\textbf{AC:} $40.4 \times 10^7$} & \multicolumn{1}{c|}{-} & \multicolumn{1}{c|}{-} & \multicolumn{1}{c|}{81.3} & - \\ \hline
\multicolumn{1}{|c|}{\multirow{2}{*}{\textit{\textbf{\begin{tabular}[c]{@{}c@{}}RSNN-BP\\ (Ours)\end{tabular}}}}} & \multicolumn{2}{c|}{\textbf{HoNB} (2000 Neurons)} & \multicolumn{1}{c|}{AC: $9.54 \times 10^7$} & \multicolumn{1}{c|}{94.87} & \multicolumn{1}{c|}{82.89} & \multicolumn{1}{c|}{80.25} & 97.06 \\ \cline{2-8} 
\multicolumn{1}{|c|}{} & \multicolumn{2}{c|}{\textbf{HeNB} (2000 Neurons)} & \multicolumn{1}{c|}{AC: $ 9.18 \times 10^7$} & \multicolumn{1}{c|}{96.84} & \multicolumn{1}{c|}{88.36} & \multicolumn{1}{c|}{84.32} & 98.12 \\ \hline
\multicolumn{8}{|c|}{\textit{\textbf{Unsupervised learning method}}} \\ \hline
\multicolumn{1}{|c|}{\multirow{2}{*}{}} & \multicolumn{1}{c|}{\multirow{2}{*}{\textbf{Model}}} & \multicolumn{1}{c|}{\multirow{2}{*}{\textbf{No of neurons}}} & \multicolumn{1}{c|}{\multirow{2}{*}{\textbf{\begin{tabular}[c]{@{}c@{}}MAC/ Avg. Neuron \\ Activation ($\bar{\nu}$)\end{tabular}}}} & \multicolumn{3}{c|}{\textbf{RGB Datasets}} & \textbf{Event Dataset} \\ \cline{5-8} 
\multicolumn{1}{|c|}{} & \multicolumn{1}{c|}{} & \multicolumn{1}{c|}{} & \multicolumn{1}{c|}{} & \multicolumn{1}{c|}{\textbf{KTH}} & \multicolumn{1}{c|}{\textbf{UCF11}} & \multicolumn{1}{c|}{\textbf{UCF101}} & \textbf{DVS128} \\ \hline
\multicolumn{1}{|c|}{\multirow{2}{*}{\textbf{\begin{tabular}[c]{@{}c@{}}DNN- \\ Unsupervised\end{tabular}}}} & \multicolumn{1}{c|}{MetaUVFS \cite{patravali2021unsupervised}}  & \multicolumn{1}{c|}{-} & \multicolumn{1}{c|}{\textbf{MAC}: $58.39 \times 10^9$} & \multicolumn{1}{c|}{90.14} & \multicolumn{1}{c|}{80.79} & \multicolumn{1}{c|}{76.38} & - \\ \cline{2-8} 
\multicolumn{1}{|c|}{} & \multicolumn{1}{c|}{Soomro et al. \cite{soomro2017unsupervised}} & \multicolumn{1}{c|}{-} & \multicolumn{1}{c|}{\textbf{MAC}: $63 \times 10^9$} & \multicolumn{1}{c|}{84.49} & \multicolumn{1}{c|}{73.38} & \multicolumn{1}{c|}{61.2} & - \\ \hline

\multicolumn{1}{|c|}{\multirow{3}{*}{\textbf{SNN-Unsupervised}}} & \multicolumn{1}{c|}{GRN-BCM \cite{meng2011modeling}} & \multicolumn{1}{c|}{1536} & \multicolumn{1}{c|}{$\mathbf{\nu} = 3.56 \times 10^3$} & \multicolumn{1}{c|}{74.4} & \multicolumn{1}{c|}{-} & \multicolumn{1}{c|}{-} & 77.19 \\ \cline{2-8} 
\multicolumn{1}{|c|}{} & \multicolumn{1}{c|}{LSM-STDP \cite{ivanov2021increasing}} & \multicolumn{1}{c|}{135} & \multicolumn{1}{c|}{$\mathbf{\nu} = 10.12 \times 10^3$} & \multicolumn{1}{c|}{66.7} & \multicolumn{1}{c|}{-} & \multicolumn{1}{c|}{-} & 67.41 \\ \cline{2-8} 
\multicolumn{1}{|c|}{} & \multicolumn{1}{c|}{\begin{tabular}[c]{@{}c@{}}GP-Assisted\\ CMA-ES  \cite{zhou2020surrogate}\end{tabular}} & \multicolumn{1}{c|}{500} & \multicolumn{1}{c|}{$\mathbf{\nu} = 9.23 \times 10^3$} & \multicolumn{1}{c|}{87.64} & \multicolumn{1}{c|}{-}  &  \multicolumn{1}{c|}{-} & 89.25 \\ \hline

\multicolumn{1}{|c|}{\multirow{3}{*}{\textit{\textbf{\begin{tabular}[c]{@{}c@{}}RSNN-STDP \\ Unsupervised\\ (Ours)\end{tabular}}}}} & \multicolumn{1}{c|}{\textbf{HoNHoS}} & \multicolumn{1}{c|}{2000} & \multicolumn{1}{c|}{$\mathbf{\nu} = 3.85 \times 10^3$} & \multicolumn{1}{c|}{86.33} & \multicolumn{1}{c|}{75.23} & \multicolumn{1}{c|}{74.45} & 90.33 \\ \cline{2-8} 
\multicolumn{1}{|c|}{} & \multicolumn{1}{c|}{\textbf{HeNHeS}} & \multicolumn{1}{c|}{500} & \multicolumn{1}{c|}{$\mathbf{\nu} = 2.93 \times 10^3$} & \multicolumn{1}{c|}{88.04} & \multicolumn{1}{c|}{71.42} & \multicolumn{1}{c|}{70.16} & 90.15 \\ \cline{2-8} 
\multicolumn{1}{|c|}{} & \multicolumn{1}{c|}{\textbf{HeNHeS}} & \multicolumn{1}{c|}{2000} & \multicolumn{1}{c|}{$\mathbf{\nu} = 2.74 \times 10^3$} & \multicolumn{1}{c|}{94.32} & \multicolumn{1}{c|}{79.58} & \multicolumn{1}{c|}{77.33} & 96.54 \\ \hline
\end{tabular}%
}
\end{table}

\section{Conclusions}
We develop a novel method using recurrent SNN to classify spatio-temporal signals for action recognition tasks using biologically-plausible unsupervised STDP learning. We show how heterogeneity in the neuron parameters and the LTP/LTD dynamics of the STDP learning process can improve the performance and empirically demonstrate their impact on developing smaller models with sparse connections and trained with lesser training data. It is well established in neuroscience that, heterogeneity \cite{de1987feedback}, \cite{petitpre2018neuronal}, \cite{shamir2006implications} is an intrinsic property of the human brain. Our analysis shows that incorporating such concepts is beneficial for designing high-performance HRSNN for classifying complex spatio-temporal datasets for action recognition tasks.




\section{Supplementary Section}

\section*{Theorem 1}

\par \textbf{ Equation for Describing State Dynamics of RSNN } We define the state update equation for the recurrent spiking neural network is given as:
\begin{align}
    \mathbf{X}(t+1)&=[\mathbf{A} \mid \mathbf{B}] \cdot[\mathbf{X} \mid \mathbf{U}]^{\top} \nonumber \\
[\mathbf{A} \mid  \mathbf{B}]&= \mathbf{X}(t+1) \cdot \left([\mathbf{X} \mid \mathbf{U}]^{\top}\right)^{\dagger} \nonumber \\
\mathbf{Z}=\mathbf{W} \cdot \mathbf{X} &\Rightarrow[\mathbf{A} \mid  \mathbf{B}]=\mathbf{Z} \cdot (\mathbf{X})^{\dagger}
\end{align}

For brevity and simplicity, in the rest of the paper, we assume the hidden state $h_{j}^{t}$ of a LIF neuron model contains only an activity value $v_{j}^{t}$ that evolves over time according to the equation
\begin{align}
    v_{j}^{t+1}&=\alpha v_{j}^{t}+\sum_{i \neq j} \hat{W}_{j i} z_{i}^{t}+\sum_{i} W_{j i}^{\mathrm{in}} x_{i}^{t+1}-z_{j}^{t} v_{\mathrm{th}} \nonumber \\
    z_{j}^{t} &=\sigma\left(v_{j}^{t}-v_{\mathrm{th}}\right)
    \label{eq:states1}
\end{align}

where $\sigma$ is the nonlinearity (e.g., the Heaviside step function), $v_{j}^{t}$ is the activity of neuron $j$ at discrete time $t$, and $v_{\text {th }}$ is the threshold constant. A neuron spikes $\left(z_{j}^{t}=1\right)$ if its activity reaches the activity threshold, and remains silent $\left(z_{j}^{t}=0\right)$ otherwise.
$W_{j i}^{\text {rec }}$ is a synapse weight from neuron $i$ to neuron $j$, and $\alpha$ is a constant decay factor. The first term in the above equation models the decay of the activity value over time. The second and third terms model the input of the neuron from other neurons or from the input to the network, respectively. The fourth term $\left(-z_{j}^{t} v_{\mathrm{th}}\right)$ ensures that the activity of the neuron drops when it spikes. Hence, we can rewrite Eq.\ref{eq:states1} as follows:
\begin{equation}
    \left(\alpha \mathbf{x}(t-1)+\sigma \left(\mathbf { W }_{in}  F  \left[\mathbf{u}(t), \mathbf{x}(t-1)\right]+\boldsymbol{\theta}+\hat{\mathbf{W}} \mathbf{x}(t-1)\right)\right)
\end{equation}

For this proof, we consider the HRSNN as a netorked dynamical system and follow a similar analysis as done by Tu et al. \cite{tu2021dimensionality}.
Let us consider a networked system consisting of $N$ nodes whose states $x=\left(x_{1}, \ldots, x_{N}\right)^{\top}$ follow the dynamic equation
\begin{equation}
    \frac{d x_{i}}{d t}=F_{i}\left(x_{i}\right)+\sum_{j}^{N} A_{i j} G_{i}\left(x_{i}, x_{j}\right)
\label{eq:1}
\end{equation}

where $F_{i}\left(x_{i}\right)$ is the "local" dynamics at node $i$ (or "self-dynamics") and $G_{i}\left(x_{i}, x_{j}\right)$ is the dynamics expressing the coupling of node $i$ with its neighbors $j$ , according to the adjacency matrix $A \in$ $R^{N \times N}$, representing the interaction network of the system, with $A_{i j}$ capturing the interaction $i \leftarrow j$. Recently, Gao et al.\cite{gao2016universal} investigated the resilience of this system in the particular case in which the functions $F$ and $G$ expressing the self-dynamics and coupling-dynamics are the same at all nodes, i.e., $\forall i, F_{i}\left(x_{i}\right)=F\left(x_{i}\right)$ and $\forall i, G_{i}\left(x_{i}, x_{j}\right)=G\left(x_{i}, x_{j}\right)$.
We define the mean field operator \cite{gao2016universal} 
$$ \mathcal{L}(\mathbf{x})=\frac{1}{N} \sum_{j=1}^{N} s_{j}^{\text {out }} x_{j} / \frac{1}{N} \sum_{j=1}^{N} s_{j}^{\text {out }}=\frac{\left\langle\mathbf{s}^{\text {out }} \cdot \mathbf{x}\right\rangle}{\left\langle\mathbf{s}^{\text {out }}\right\rangle}$$ where $\mathbf{s}^{\text {out }}=\left(s_{1}^{\text {out }}, \ldots, s_{N}^{\text {out }}\right)$ is the vector of the out-degree of matrix $A$; then, we characterize the effective state of the networked system using the weighted average node state $x_{e f f}=\mathcal{L}(x)$. If the network's degree correlation is low, we can assume that the Hadamard product approximation holds. Then, applying Chebyshev expansion to approximate $F_{i}\left(x_{i}\right)$ and $G_{i}\left(x_{i}, x_{j}\right)$ with polynomial functions of order $m$ and $n$, respectively, Equation \ref{eq:1} can be reduced to
\begin{equation}
    I\left(d_{1}, \ldots, d_{s}, x_{\text {eff }}\right)=\frac{d x_{\text {eff }}}{d t} \approx \sum_{s=1}^{s} d_{s} * x_{\text {eff }}^{s-1}
    \label{eq:2}
\end{equation}

where $S=\max (m, n), d_{s}=\{\begin{array}{l}B_{\text {eff }}^{s}+A_{\text {eff }} * C_{\text {eff }}^{s}, s \in[1, \min (m, n)] \\ A_{\text {eff }}^{s} C_{\text {eff }}^{s}, s \in[m+1, n], m<n \\ B_{\text {eff }}^{s}, s \in[n+1, m], n<m\end{array}$ ; 

$A_{\text {eff }}=\mathcal{L}\left(s^{i n}\right), B_{\text {eff }}^{s}=\mathcal{L}\left(B^{s}\right).$, and $C_{\text {eff }}^{s}=\mathcal{L}\left(C^{s}\right)$. $B^{k}=\left(b_{1, k}, \ldots, b_{N, k}\right)^{\top}$ is the column of the $k$-th term of the $m$-order Chebyshev polynomials approximating the self-dynamics $F_{i}\left(x_{i}\right)$, and $C^{\prime}=\left(c_{1, l}, \ldots, c_{N, l}\right)^{T}$ is the column of the l-th factor of the n-order Chebyshev polynomials approximating the coupling-dynamics $G_{i}\left(x_{i}, x_{j}\right)$. Therefore, we map the dynamics of Equation \ref{eq:1}
into Equation\ref{eq:2} and study the resilience of the system, through the behavior of $x_{eff}$ at steady state and
its response to a perturbation of one or more of these $S$ parameters. In particular, the conditions for stability of a state $x_{\text {eff }}^{*}$ of the dynamics can thus be associated with a region expressed by the equation set:
$$\left\{\begin{array}{l}I\left(d_{1}, \ldots, d_{s}, x_{\text {eff }}^{*}\right)=0 \\ \frac{d l}{d x_{e f f}}<0\end{array}\right.$$ 
where the function $I$ represents the system's dynamics and $d_{1}, \ldots, d_{S}$ are their control parameters.

Now, for homogeneous and heterogeneous RSNNs, the polynomial approximations using polynomial chaos is derived by Kubota et al. \cite{kubota2021unifying}.

\textit{\textbf{Theorem 1: }  Assuming $\mathcal{S}_u$ is finite and contains $s$ inputs, let $r_{\text{Hom}}, r_{\text{Het}}$ are the ranks of the $n \times s$ matrices consisting of the $s$ vectors $\mathbf{x}_{u}\left(t_{0}\right)$ for all inputs $u$ in $\mathcal{S}_u$ for each of Homogeneous and Heterogeneous RSNNs respectively. Then $r_{\text{Hom}} \leq r_{\text{Het}}$.}

\textbf{Proof: }
To prove that the rank of the Heterogeneous state matrix is greater than the rank of the homogeneous one, we aim to show that the number of linearly independent vectors for HeNHeS is greater than or equal to the number of linearly independent vectors for HoNHoS. However, since the state-space of a heterogeneous network is very high dimensional, we aim to show the results for a low-dimensional projection of this high-dimensional hyperspace. In other words, we aim to show that the  number of dimensions of a low-rank approximation of the state-space of HeNHeS model is greater than the HoNHoS model. For this proof, we treat the HRSNN as a use a heterogeneous graph and use the network representation learning framework in order to embed the network nodes into a low-dimensional vector space, by preserving network topology structure, node and edge information.

First, let us consider that the response of neuron $i \in \mathcal{R}$ is given as
\begin{equation}
    \mathbf{y}_{i}=x^{(1)} \boldsymbol{\beta}_{i}^{1}+\ldots x^{(N)} \boldsymbol{\beta}_{i}^{N}+\mathbf{b}_{i}+\boldsymbol{\epsilon}_{i}
\end{equation}
where $\mathbf{b}_{i}$ is a constant vector representing a condition-independent mean, and $\boldsymbol{\epsilon}_{i}$ is noise. The state-space description of the response is represented by a factorization of the vectors $\boldsymbol{\beta}_{i}^{ \top}=\mathbf{S}^{\top} \mathbf{w}_{i}$ where, $r$ is the dimensionality of the subspace for states of the neurons in $\mathcal{R}$. Thus, $\mathbf{w}_{i} \in \mathbf{R}^{r}$ is a neuron-specific vector of weights and $\mathbf{S}$ is a matrix of rank $r$. If $\mathbf{w}_{i}^{\top}=\left(\mathbf{w}_{i}^{1 \top}, \ldots, \mathbf{w}_{i}^{\top}\right)$, and $\mathbf{S}$ be a block-diagonal matrix given as follows:
\begin{equation}
   \mathbf{S}=\left(\begin{array}{ccc}
\mathbf{S}_{1} & & \\
& \ddots & \\
& & \mathbf{S}_{P}
\end{array}\right)
\end{equation}

then we get
\begin{equation}
    \mathbf{y}_{i}=\left(\mathbf{x}^{\top} \otimes I_{T}\right) \mathbf{S}^{\top} \mathbf{w}_{i}+\mathbf{b}_{i}+\boldsymbol{\epsilon}_{i} 
\end{equation}

If $\mathbf{y}_{i}$ and $\mathbf{x}$ are the observed response and states of the recurrent neurons, then the collection of all observations for this neuron $\mathbf{y}_{i}^{\top}=\left(\mathbf{y}_{i, 1}^{\top}, \ldots, \mathbf{y}_{i, N}^{\top}\right)$ can be described in terms of all the neuron states in $\mathcal{R}$: $\mathbf{X}_{i}^{\top}=\left(\mathbf{x}_{1}, \ldots, \mathbf{x}_{N}\right)$ by

\begin{align}
\mathbf{y}_{i} &=\left(\mathbf{X}_{i} \otimes I_{T}\right) \mathbf{S}^{\top} \mathbf{w}_{i}+\mathbf{1}_{N} \otimes \mathbf{b}_{i}+\boldsymbol{\epsilon}_{i} \\
&=\mathbf{F}_{i} \mathbf{w}_{i}+\mathbf{b}_{i}^{\prime}+\boldsymbol{\epsilon}_{i}
\label{eq:5}
\end{align}

where $\mathbf{F}_{i}=\left(\mathbf{X}_{i}^{\top} \otimes I_{T}\right) \mathbf{S}^{\top}, \mathbf{b}_{i}^{\prime}=\mathbf{1}_{N} \otimes \mathbf{b}_{i}$, where $\boldsymbol{\epsilon}_{i}^{\top}=$ $\left(\boldsymbol{\epsilon}_{i, 1}^{\top}, \ldots, \boldsymbol{\epsilon}_{i, N}^{\top}\right)$.



The rank of the model corresponds to the rank of  $\mathbf{B}$. We first estimate the model parameters with rank $r=0$ denoting the null model for all elements of $\mathbf{B}$. For HeNHeS model, the variance of the

Again, let us fix some inputs $u_{1}, \ldots, u_{r}$ in $S_{\text {univ }}$ so that the resulting $r$ circuit states $\mathbf{x}_{u_{i}}\left(t_{0}\right)$ are linearly independent. The rather small rank of the state matrix, especially in the ordered regime, can be explained in part by the small number of neurons which get activated (i.e., emit at least one spike) for a given input pattern. For some input pattern $u$, let the activation vector $\mathbf{x}_{u}^{\text {act }} \in\{0,1\}^{n}$ be the vector with the $i$ th entry being 1 if neuron $i$ was activated during the presentation of this pattern. Thus, for HRSNN with HeNHeS, the number of neurons which get activated are higher than HoNHoS models. Hence $r_{\text{Hom}} \le r_{\text{Het}}$.


\section*{Theorem 2:}

Gaussian processes are a useful technique for modeling unknown functions. We study how to extend this model class to model functions in a Wasserstein metric space. We do so in a manner which is both mathematically well-posed, and constructive enough to allow the kernel to be computed. This allows the said processes to be trained with standard methods and also enables their use in Bayesian optimization of the hyperparameters of the RSNN.

\par \textbf{Joint Probability Distribution:} We consider the Wasserstein Distance between the joint probability distributions of all the distributions of all the hyperparameters used. The histogram of the hyperparameters which has heterogeneity in their parameters shows a distribution of the parameters. For fixed hyperparameters which are also tuned, we consider a Delta Dirac Distribution at the value of the hyperparameters.

\par \textbf{Matern Kernel: } One of the most widely-used kernels is the Matérn kernel, which is given by
$$
\mathcal{K}\left(x, x^{\prime}\right)=\sigma^{2} \frac{2^{1-\nu}}{\Gamma(\nu)}\left(\sqrt{2 \nu} \frac{\left\|x-x^{\prime}\right\|}{\kappa}\right)^{\nu} K_{\nu}\left(\sqrt{2 \nu} \frac{\left\|x-x^{\prime}\right\|}{\kappa}\right)
$$
where $K_{\nu}$ is the modified Bessel function of the second kind, and $\sigma^{2}, \kappa, \nu$ are the variance, length scale, and smoothness parameters, respectively.

\par \textbf{Wasserstein Metric Space: }

Let $\sigma$ and $\mu$ be two probability measures on measurable spaces $X$ and $Y$ and their corresponding probability density functions $I_{0}$ and $I_{1}, d \sigma(x)=I_{0}(x) d x$ and $d \mu(y)=I_{1}(y) d y$. 

\textbf{Definition} The $p$-Wasserstein distance for $p \in[1, \infty)$ is defined as,
$$
W_{p}(\sigma, \mu):=\left(\inf _{\pi \in \Pi(\sigma, \mu)} \int_{X \times Y}(x-y)^{p} d \pi(x, y)\right)^{\frac{1}{p}}
$$
where $\Pi(\sigma, \mu)$ is the set of all transportation plans, and $\pi \in$ $\Pi(\sigma, \mu)$ such that 
$\pi(A \times Y)=\sigma(A) \quad$ for any Borel subset $A \subseteq X$ and,
$\pi(X \times B)=\mu(B) \quad$ for any Borel subset $B \subseteq Y$.
Using Brenier's theorem, for absolutely continuous probability measures $\sigma$ and $\mu$ with respect to Lebesgue measure, the $p$-Wasserstein distance can be derived as,
$$
W_{p}(\sigma, \mu)=\left(\inf _{f \in M P(\sigma, \mu)} \int_{X}(f(x)-x)^{p} d \sigma(x)\right)^{\frac{1}{p}}
$$
where, $M P(\sigma, \mu)=\left\{f: X \rightarrow Y \mid f_{\#} \sigma=\mu\right\}$ and $f_{\#} \sigma$ represents the pushforward of measure $\sigma$ and is characterized as,
$\int_{f^{-1}(A)} d \sigma=\int_{A} d \mu$ for any Borel subset $A \subseteq Y$

\par \textbf{Sliced Wasserstein Distance: } We use the sliced Wasserstein distance to represent the family of one-dimensional distributions for the higher-dimensional probability distribution, and then calculate the distance between two input higher-dimensional distributions as a functional on the Wasserstein distance of their one-dimensional representations. In this sense, the distance is obtained by solving several one-dimensional optimal transport problems, which have closed-form solutions.

\textbf{Definition} Let $\sigma$ and $\mu$ be two continuous probability measures on $\mathbb{R}^{d}$ with corresponding positive probability density functions $I_{1}$ and $I_{0}$. The Sliced Wasserstein distance between $\mu$ and $\sigma$ is defined as,
$$
\begin{aligned}
W_S(\mu, \sigma) &:=\left(\int_{\mathbb{S}^{d-1}} W_{2}^{2}\left(\mathcal{S} I_{1}(., \theta), \mathcal{S} I_{0}(., \theta)\right) d \theta\right)^{\frac{1}{2}} \\
&=\left(\int_{\mathbb{S}^{d-1}} \int_{\mathbb{R}}\left(f_{\theta}(t)-t\right)^{2} \mathcal{S} I_{0}(t, \theta) d t d \theta\right)^{\frac{1}{2}}
\end{aligned}
$$
where $f_{\theta}$ is the MP map between $\mathcal{S} I_{0}(., \theta)$ and $\mathcal{S} I_{1}(., \theta)$ such that,
$$
\int_{-\infty}^{f_{\theta}(t)} \mathcal{S} I_{1}(\tau, \theta) d \tau=\int_{-\infty}^{t} \mathcal{S} I_{0}(\tau, \theta) d \tau, \forall \theta \in \mathbb{S}^{d-1}
$$
or equivalently in the differential form,
$$
\frac{\partial f_{\theta}(t)}{\partial t} \mathcal{S} I_{1}\left(f_{\theta}(t), \theta\right)=\mathcal{S} I_{0}(t, \theta), \quad \forall \theta \in \mathbb{S}^{d-1} .
$$

\par \textbf{Positive Definite:}  A positive definite (PD) (resp. conditional negative definite) kernel on a set $M$ is a symmetric function $\mathcal{K}: M \times M \rightarrow \mathbb{R}, \mathcal{K}\left(I_{i}, I_{j}\right)=\mathcal{K}\left(I_{j}, I_{i}\right)$ for all $I_{i}, I_{j} \in M$, such that for any $n \in N$, any elements $I_{1}, \ldots, I_{n} \in X$, and numbers $c_{1}, \ldots, c_{n} \in \mathbb{R}$, we have
$$
\sum_{i=1}^{n} \sum_{j=1}^{n} c_{i} c_{j} \mathcal{K}\left(I_{i}, I_{j}\right) \geq 0 \quad(\text { resp. } \leq 0)
$$
with the additional constraint of $\sum_{i=1}^{n} c_{i}=0$ for the conditionally negative definiteness.

We first demonstrate that the Sliced Wasserstein Matern kernel of probability measures is a positive definite kernel. We proceed our argument by showing that there is an explicit formulation for the nonlinear mapping to the kernel space and define a family of kernels based on this mapping. We start by proving that for one-dimensional probability density functions the 2-Wasserstein Matern kernel is a positive definite kernel.

First, we start by proving that for one-dimensional probability density functions the 2-Wasserstein Matern kernel is a positive definite kernel.

\par \textit{\textbf{Theorem :} Let $M$ be the set of absolutely continuous one-dimensional positive probability density functions and define $\mathcal{K}: M \times M \rightarrow \mathbb{R}$ to be
$\mathcal{K}\left(I_{i}, I_{j}\right):= \sigma^{2} \frac{2^{1-\nu}}{\Gamma(\nu)}\left(\sqrt{2 \nu} \frac{W_{2}\left(I_{i}, I_{j}\right)}{\kappa}\right)^{\nu} K_{\nu}\left(\sqrt{2 \nu} \frac{W_{2}\left(I_{i}, I_{j}\right)}{\kappa}\right)$, then $\mathcal{K}(., .)$ is a positive definite kernel for all $\gamma>0$.}

Here, $K_{\nu}$ is the modified Bessel function of the second kind, and $\sigma^{2}, \kappa, \nu$ are the variance, length scale, and smoothness parameters, respectively. 

\textbf{Proof: } In order to be able to show this, we first show that for absolutely continuous one-dimensional positive probability density functions there exists an inner product space $\mathcal{V}$ and a function $\psi: M \rightarrow \mathcal{V}$ such that $W_{2}\left(I_{i}, I_{j}\right)=\| \psi\left(I_{i}\right)-$ $\psi\left(I_{j}\right) \|_\mathcal{V}$

Let $\sigma, \mu$, and $\nu$ be probability measures on $\mathbb{R}$ with corresponding absolutely continuous positive density functions $I_{0}, I_{1}$, and $I_{2}$. Let $f, g, h: \mathbb{R} \rightarrow \mathbb{R}$ be transport maps such that $f_{\#} \sigma=\mu, g_{\#} \sigma=\nu$, and $h_{\#} \mu=\nu$. In the differential form this is equivalent to $f^{\prime} I_{1}(f)=g^{\prime} I_{2}(g)=I_{0}$ and $h^{\prime} I_{2}(h)=I_{1}$ where $I_{1}(f)$ represents $I_{1} \circ f$. Then we have,
$$
\begin{aligned}
W_{2}\left(I_{1}, I_{0}\right) &=\int_{\mathbb{R}}(f(x)-x) I_{0}(x) d x \\
W_{2}\left(I_{2}, I_{0}\right) &=\int_{\mathbb{R}}(g(x)-x) I_{0}(x) d x \\
W_{2}\left(I_{2}, I_{1}\right) &=\int_{\mathbb{R}}(h(x)-x) I_{1}(x) d x
\end{aligned}
$$
We follow the work of Wang et al. [45] and Park et al. [31] and define a nonlinear map with respect to a fixed probability measure, $\sigma$ with corresponding density $I_{0}$, that maps an input probability density to a linear functional on the corresponding transport map. More precisely, $\psi_{\sigma}\left(I_{1}(.)\right):=$ $(f(.)-i d(.)) \sqrt{I_{0}(.)}$ where $i d(.)$ is the identity map and $f^{\prime} I_{1}(f)=I_{0}$. Notice that such $\psi_{\sigma}$ maps the fixed probability density $I_{0}$ to zero, $\psi_{\sigma}\left(I_{0}(.)\right)=(i d(.)-i d(.)) \sqrt{I_{0}(.)}=$ 0 and it satisfies,
$$
\begin{aligned}
&W_{2}\left(I_{1}, I_{0}\right)=\left\|\psi_{\sigma}\left(I_{1}\right)\right\|_{2} \\
&W_{2}\left(I_{2}, I_{0}\right)=\left\|\psi_{\sigma}\left(I_{2}\right)\right\|_{2}
\end{aligned}
$$
More importantly, we demonstrate that $W_{2}\left(I_{2}, I_{1}\right)=$ $\left\|\psi_{\sigma}\left(I_{1}\right)-\psi_{\sigma}\left(I_{2}\right)\right\|_{2}$. To show this, we can write,
$$
\begin{aligned}
W_{2}\left(I_{2}, I_{1}\right) &=\int_{\mathbb{R}}(h(x)-x) I_{1}(x) d x \\
&=\int_{\mathbb{R}}(h(f(\tau))-f(\tau)) f^{\prime}(\tau) I_{1}(f(\tau)) d \tau \\
&=\int_{\mathbb{R}}(g(\tau)-f(\tau)) I_{0}(\tau) d \tau \\
&=\int_{\mathbb{R}}((g(\tau)-\tau)-(f(\tau)-\tau)) I_{0}(\tau) d \tau \\
&=\left\|\psi_{\sigma}\left(I_{1}\right)-\psi_{\sigma}\left(I_{2}\right)\right\|_{2}
\end{aligned}
$$
Finally, we know that the one-dimensional transport maps are unique, therefore if $(h \circ f) \# \sigma=\nu$ and $g_{\#} \sigma=\nu$ then $h \circ f=g .$

We showed that there exists a nonlinear map $\psi_{\sigma}: M \rightarrow$ $\mathcal{V}$ for which $W_{2}\left(I_{i}, I_{j}\right) = \left\|\psi_{\sigma}\left(I_{i}\right) - \psi_{\sigma}\left(I_{j}\right)\right\|_{2}$ and as shown by Jayasumana et al. \cite{jayasumana2015kernel} and Kolouri et al. \cite{kolouri2016sliced}, we can conclude that, $\mathcal{K}\left(I_{i}, I_{j}\right)$ is a positive definite kernel.

\textit{\textbf{Theorem 2:} The modified Matern function on the Wasserstein metric space $\mathcal{W}$ is a valid kernel function}

\textbf{ Proof: } To show that the above function is a kernel function, we need to prove that Mercer's theorem holds. i.e., (i) the function is symmetric and (ii) in a finite input space, the Gram matrix of the kernel function is positive semi-definite. 
The Sliced Wasserstein distance as defined above is symmetric, and it satisfies subadditivity and coincidence axioms, and hence it is a true metric. \cite{kolouri2015radon}.

First note that for an absolutely continuous positive probability density function, $I \in M$, each hyperplane integral, $\mathcal{S} I(., \theta), \forall \theta \in \mathbb{S}^{d-1}$ is a one dimensional absolutely continuous positive probability density function. Therefore,
$$
\sum_{i=1}^{N} \sum_{j=1}^{N} c_{i} c_{j} W_{2}^{2}\left(\mathcal{S} I_{i}(., \theta), \mathcal{S} I_{j}(., \theta)\right) \leq 0, \forall \theta \in \mathbb{S}^{d-1}
$$
where $\sum_{i=1}^{N} c_{i}=0$. Integrating the left hand side of above inequality over $\theta$ leads to,
$$
\begin{gathered}
\int_{\mathbb{S}^{d-1}}\left(\sum_{i=1}^{N} \sum_{j=1}^{N} c_{i} c_{j} W_{2}^{2}\left(\mathcal{S} I_{i}(., \theta), \mathcal{S} I_{j}(., \theta)\right) d \theta\right) \leq 0 \Rightarrow \\
\sum_{i=1}^{N} \sum_{j=1}^{N} c_{i} c_{j}\left(\int_{\mathbb{S}^{d-1}} W_{2}^{2}\left(\mathcal{S} I_{i}(., \theta), \mathcal{S} I_{j}(., \theta)\right) d \theta\right) \leq 0 \Rightarrow \\
\sum_{i=1}^{N} \sum_{j=1}^{N} c_{i} c_{j} W_{S}^{2}\left(I_{i}, I_{j}\right) \leq 0
\end{gathered}
$$
Therefore $W_{S}^{2}(., .)$ is conditionally negative definite, and hence from the previous theorem we have that $\mathcal{K}\left(I_{i}, I_{j}\right)$  is a positive definite kernel for $\gamma>$ 0 .

\section*{Bayesian Optimization}
\begin{table}[b]
\centering
\caption{Table showing the notations used in the paper}
\label{tab:not}
\resizebox{\columnwidth}{!}{%
\begin{tabular}{|c|c|c|c|}
\hline
\textbf{Notation} & \textbf{Full Form}                & \textbf{Notation}           & \textbf{Full Form}                 \\ \hline
\textbf{SNN}      & Spiking Neural Network            & $\mathbf{\mathcal{I}}$     & Input Layer                        \\ \hline
\textbf{RSNN}     & Recurrent SNN                     & $\mathbf{\mathcal{R}} $     & Recurrent Layer                    \\ \hline
\textbf{HRSNN}    & Heterogeneous RSNN                & $\mathbf{\mathcal{O}}$      & Output Layer                       \\ \hline
\textbf{STDP}     & Spike Timing Dependent Plasticity & $\mathbf{\mathcal{S}_{XY}}$ & Connections between layers X and Y \\ \hline
\textbf{LIF}      & Leaky Integrate and Fire          & $\mathbf{N} $               & Number of neurons                  \\ \hline
\textbf{HoNHoS} & Homogeneous LIF, Homogeneous STDP & \textbf{HeNHeS} & Heterogeneous LIF,  Heterogeneous STDP \\ \hline
\textbf{HeNHeS} & Heterogeneous LIF, Homogeneous STDP & \textbf{HoNHeS} & Homogeneous LIF, Heterogeneous STDP \\ \hline
\end{tabular}%
}
\end{table}

\subsection*{Brain Inspired Initialization}

\begin{figure}
    \centering
    \includegraphics[width = 0.7\columnwidth]{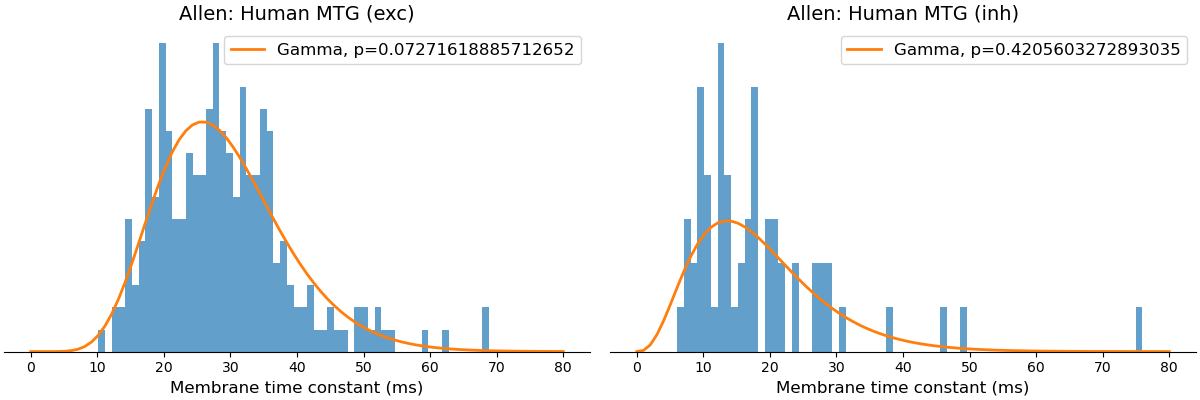}
    \caption{The fitted gamma distribution to the Allen Human brain atlas based distribution for membrane time constants}
    \label{fig:experiment_fit}
\end{figure}

Mejias et al. \cite{mejias2014differential}, showed that in real cortical populations, excitatory, and inhibitory subpopulations of neurons exhibit different cell-to-cell heterogeneities for each type of subpopulation in the system. The authors discussed the highly differentiated roles for heterogeneity, depending on excitatory or inhibitory neuron subpopulation. For example, heterogeneity among excitatory neurons non-linearly increases the mean firing rate and linearizes the f-I curves while heterogeneity among inhibitory neurons decreases the network activity level and induces divisive gain effects in the f-I curves of the excitatory cells, providing an effective gain control mechanism to influence information flow.
We use the Allen human brain-based initialization using separate distributions for the excitatory and inhibitory neuron populations. A gamma distribution is fitted using the Kernel Density Estimation Method on the data for the membrane timescales which is used to sample the values of all the membrane time constants of the recurrent neurons in the HRSNN model. The fitted distribution is shown in Fig. \ref{fig:experiment_fit}

\subsection*{Hyperparameters Optimized}

The list of the hyperparameters optimized using the Bayesian Optimization technique is shown in Table \ref{tab:BO_params}. We also show the range of the hyperparameters used and the initial values.

\begin{table}[]
\centering
\caption{The list of parameter settings for the Bayesian Optimization-based hyperparameter search}
\label{tab:BO_params}
\resizebox{0.5\textwidth}{!}{%
\begin{tabular}{|c|c|c|}
\hline
\textbf{Parameter} & \textbf{Initial Value} & \textbf{Range} \\ \hline
    $\eta$      &      10         &   (0,50)    \\ \hline
    $\gamma$      &     5          &   (0,10)    \\ \hline
     $\zeta$     &      2.5         &    (0,10)   \\ \hline
   $\eta^*$       &       1        &    (0,3)   \\ \hline
     $g$     &       2        &  (0,10)     \\ \hline
     $\omega$     &      0.5         &   (0,1)    \\ \hline
     $k$     &       50        &   (0,100)    \\ \hline
    $\lambda$ (KTH, DVS)      &        1       &    (0,2)   \\ \hline
    $\lambda$ (UCF)      &        1.5       &    (0,4)   \\ \hline
    $P_{IR}$      &      0.05         &   (0,0.1)    \\ \hline
     $\tau_{n-E}, \tau_{n-I}$ (KTH, DVS)    & $50ms$               &   $(0ms, 100ms)$    \\ \hline
     $\tau_{n-E}, \tau_{n-I}$ (UCF)    & $100ms$               &   $(0ms, 300ms)$    \\ \hline
    $A_{en-R} , A_{EE}, A_{EI}, A_{IE}, A_{II}$     &       30        &   (0,60)    \\ \hline
\end{tabular}%
}
\end{table}

\section*{Variation of rank with sparsity and weight scale}

\begin{figure}
    \centering
    \includegraphics[width=\linewidth]{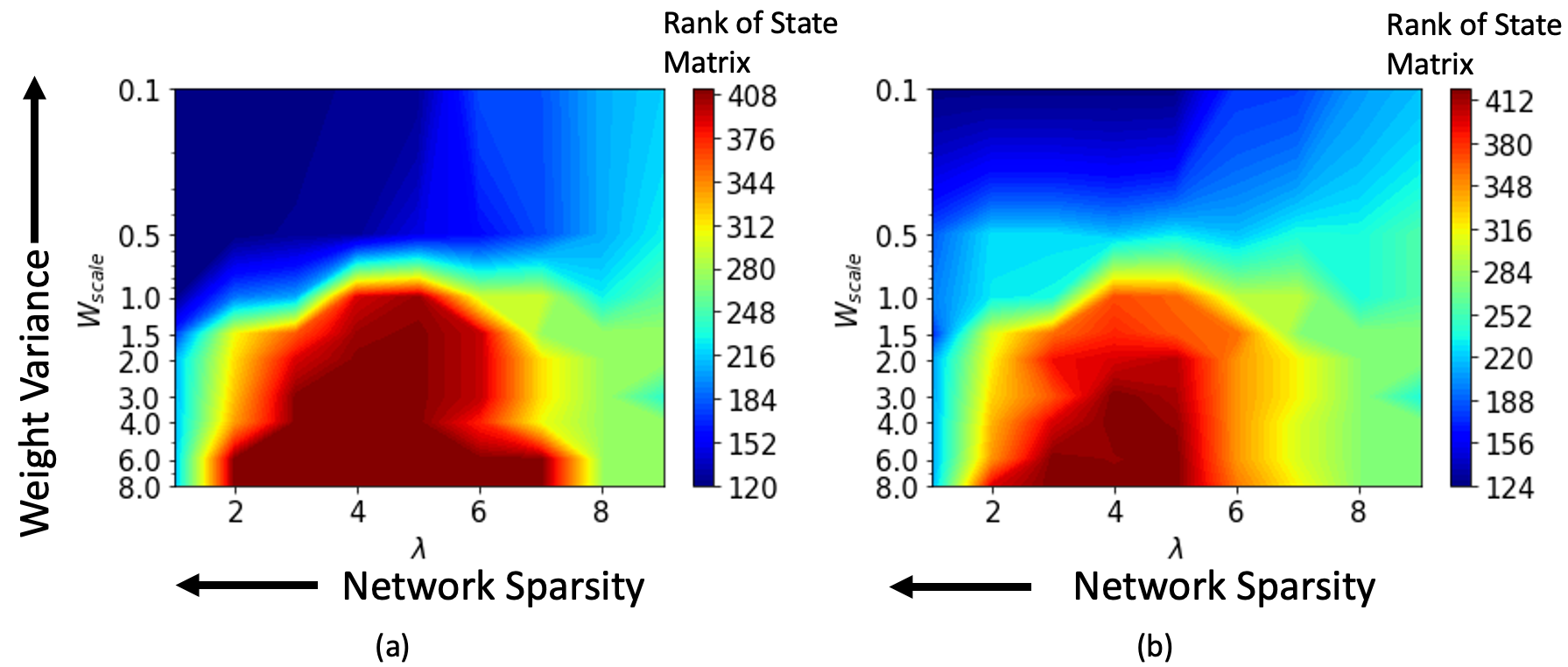}
    \caption{Change in the effective ranks of the final state matrix of HRSNN with 2000 neurons with network sparsity $\lambda$ and weight variance $W_{\text{scale}}$, for (a)HoNHoS and (b)HeNHeS. The plot is obtained by interpolating 81 points, and each point is calculated by averaging the results from 5 randomly initialized HRSNNs.}
    \label{fig:2}
\end{figure}

Here, we show the variation of the rank with the network sparsity factor $\lambda$ and the synaptic weight scale factor $W_{\text{scale}}$. The figure is shown in Fig. \ref{fig:2}. From the figure we observe the variation of the rank of the matrix with the network sparsity. This also supports our intial claim that the rank of the final state matrix can be used as a measure for the linear separation property of the HRSNN model. Comparing Figs.\ref{fig:2}(a) and (b) we also see that the performance of the model is the highest near the regions between the chaos and order. This is built on the works done by Legenstein et al. \cite{legenstein2007edge}.

\begin{figure}[b]
    \centering
    \includegraphics[width=0.3\columnwidth]{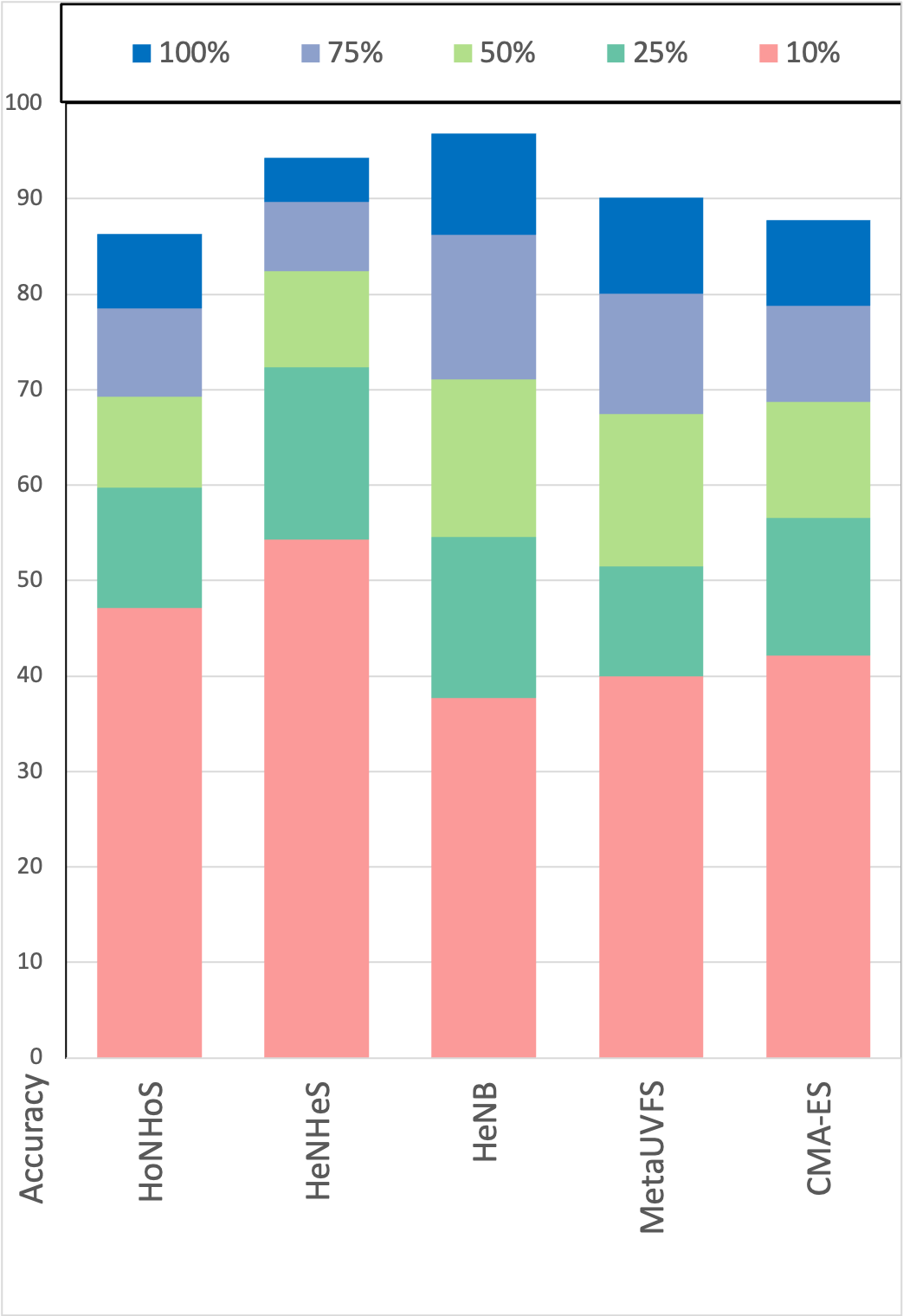}
    \caption{Figure showing the input processing and model training for the different models}
    \label{fig:reb_block}
\end{figure}

\section*{Results with Limited Training Data}

In this section, we plot the stacked bar graph for the results obtained from the DVS gesture dataset trained with limited training data. The results show a similar trend to the KTH dataset results shown in the paper.

\begin{figure}
    \centering
    \includegraphics[width=\columnwidth]{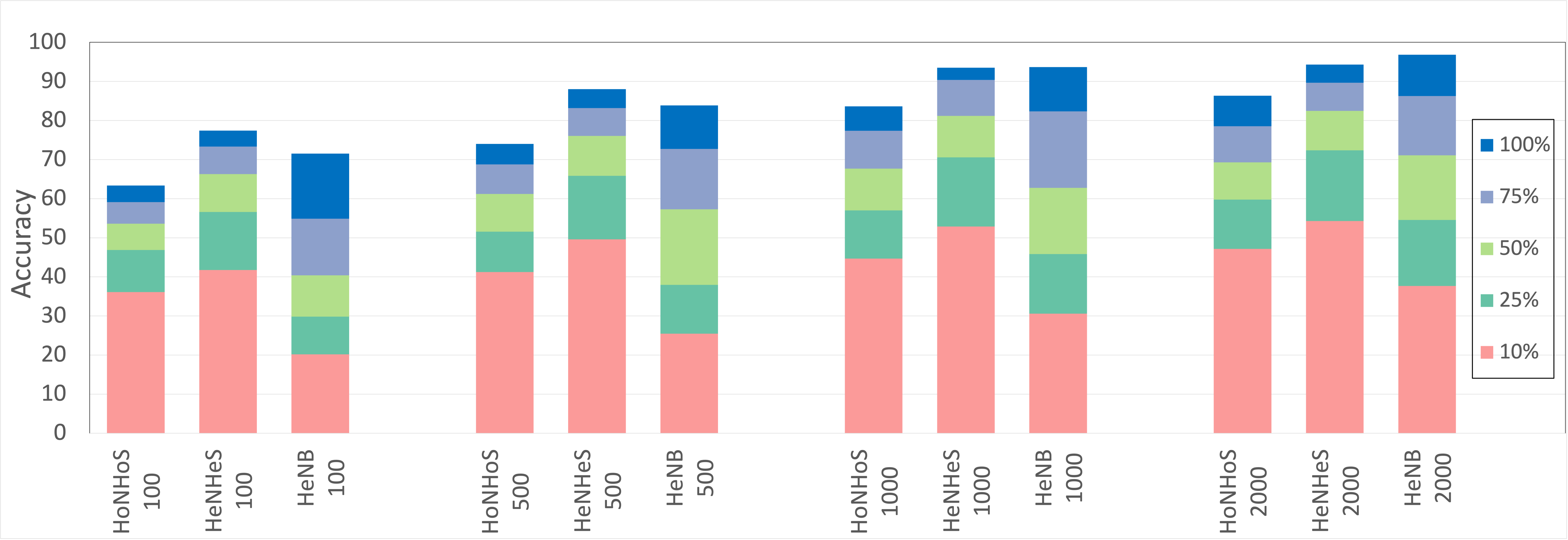}
    \caption{Bar graph showing difference in performance for the different models with
increasing training data for the DVS dataset.}
    \label{fig:my_label}
\end{figure}

\bibliographystyle{unsrt}
\bibliography{ref}

\begin{thebibliography}{10}

\bibitem{wu2018spatio}
Yujie Wu, Lei Deng, Guoqi Li, Jun Zhu, and Luping Shi.
\newblock Spatio-temporal backpropagation for training high-performance spiking
  neural networks.
\newblock {\em Frontiers in neuroscience}, 12:331, 2018.

\bibitem{jin2018hybrid}
Yingyezhe Jin, Wenrui Zhang, and Peng Li.
\newblock Hybrid macro/micro level backpropagation for training deep spiking
  neural networks.
\newblock {\em Advances in neural information processing systems}, 31, 2018.

\bibitem{shrestha2018slayer}
Sumit~B Shrestha and Garrick Orchard.
\newblock Slayer: Spike layer error reassignment in time.
\newblock {\em Advances in neural information processing systems}, 31, 2018.

\bibitem{lobo2020spiking}
Jesus~L Lobo, Javier Del~Ser, Albert Bifet, and Nikola Kasabov.
\newblock Spiking neural networks and online learning: An overview and
  perspectives.
\newblock {\em Neural Networks}, 121:88--100, 2020.

\bibitem{tavanaei2019deep}
Amirhossein Tavanaei, Masoud Ghodrati, Saeed~Reza Kheradpisheh, Timoth{\'e}e
  Masquelier, and Anthony Maida.
\newblock Deep learning in spiking neural networks.
\newblock {\em Neural Networks}, 111:47--63, 2019.

\bibitem{lazar2006combination}
Andreea Lazar, Gordon Pipa, and Jochen Triesch.
\newblock The combination of stdp and intrinsic plasticity yields complex
  dynamics in recurrent spiking networks.
\newblock In {\em ESANN}, pages 647--652. Citeseer, 2006.

\bibitem{perez2021neural}
Nicolas Perez-Nieves, Vincent~CH Leung, Pier~Luigi Dragotti, and Dan~FM
  Goodman.
\newblock Neural heterogeneity promotes robust learning.
\newblock {\em bioRxiv}, pages 2020--12, 2021.

\bibitem{she2021sequence}
Xueyuan She, Saurabh Dash, and Saibal Mukhopadhyay.
\newblock Sequence approximation using feedforward spiking neural network for
  spatiotemporal learning: Theory and optimization methods.
\newblock In {\em International Conference on Learning Representations}, 2021.

\bibitem{yin2021accurate}
Bojian Yin, Federico Corradi, and Sander~M Bohte.
\newblock Accurate online training of dynamical spiking neural networks through
  forward propagation through time.
\newblock {\em arXiv preprint arXiv:2112.11231}, 2021.

\bibitem{zeldenrust2021efficient}
Fleur Zeldenrust, Boris Gutkin, and Sophie Den{\'e}ve.
\newblock Efficient and robust coding in heterogeneous recurrent networks.
\newblock {\em PLoS computational biology}, 17(4):e1008673, 2021.

\bibitem{demin2018recurrent}
Vyacheslav Demin and Dmitry Nekhaev.
\newblock Recurrent spiking neural network learning based on a competitive
  maximization of neuronal activity.
\newblock {\em Frontiers in neuroinformatics}, page~79, 2018.

\bibitem{zhang2019spike}
Wenrui Zhang and Peng Li.
\newblock Spike-train level backpropagation for training deep recurrent spiking
  neural networks.
\newblock {\em Advances in neural information processing systems}, 32, 2019.

\bibitem{wang2021recurrent}
Zijian Wang, Yanting Zhang, Haibo Shi, Lei Cao, Cairong Yan, and Guangwei Xu.
\newblock Recurrent spiking neural network with dynamic presynaptic currents
  based on backpropagation.
\newblock {\em International Journal of Intelligent Systems}, 2021.

\bibitem{chakraborty2021characterization}
Biswadeep Chakraborty and Saibal Mukhopadhyay.
\newblock Characterization of generalizability of spike time dependent
  plasticity trained spiking neural networks.
\newblock {\em arXiv preprint arXiv:2105.14677}, 2021.

\bibitem{gilson2010stdp}
Matthieu Gilson, Anthony Burkitt, and Leo~J Van~Hemmen.
\newblock Stdp in recurrent neuronal networks.
\newblock {\em Frontiers in computational neuroscience}, 4:23, 2010.

\bibitem{nobukawa2019pattern}
Sou Nobukawa, Haruhiko Nishimura, and Teruya Yamanishi.
\newblock Pattern classification by spiking neural networks combining
  self-organized and reward-related spike-timing-dependent plasticity.
\newblock {\em Journal of Artificial Intelligence and Soft Computing Research},
  9(4):283--291, 2019.

\bibitem{ivanov2021increasing}
Vladimir Ivanov and Konstantinos Michmizos.
\newblock Increasing liquid state machine performance with edge-of-chaos
  dynamics organized by astrocyte-modulated plasticity.
\newblock {\em Advances in Neural Information Processing Systems}, 34, 2021.

\bibitem{maro2020event}
Jean-Matthieu Maro, Sio-Hoi Ieng, and Ryad Benosman.
\newblock Event-based gesture recognition with dynamic background suppression
  using smartphone computational capabilities.
\newblock {\em Frontiers in neuroscience}, 14:275, 2020.

\bibitem{lagorce2016hots}
Xavier Lagorce, Garrick Orchard, Francesco Galluppi, Bertram~E Shi, and Ryad~B
  Benosman.
\newblock Hots: a hierarchy of event-based time-surfaces for pattern
  recognition.
\newblock {\em IEEE transactions on pattern analysis and machine intelligence},
  39(7):1346--1359, 2016.

\bibitem{george2020reservoir}
Arun~M George, Dighanchal Banerjee, Sounak Dey, Arijit Mukherjee, and
  P~Balamurali.
\newblock A reservoir-based convolutional spiking neural network for gesture
  recognition from dvs input.
\newblock In {\em 2020 International Joint Conference on Neural Networks
  (IJCNN)}, pages 1--9. IEEE, 2020.

\bibitem{zhou2020surrogate}
Yan Zhou, Yaochu Jin, and Jinliang Ding.
\newblock Surrogate-assisted evolutionary search of spiking neural
  architectures in liquid state machines.
\newblock {\em Neurocomputing}, 406:12--23, 2020.

\bibitem{panda2018learning}
Priyadarshini Panda and Narayan Srinivasa.
\newblock Learning to recognize actions from limited training examples using a
  recurrent spiking neural model.
\newblock {\em Frontiers in neuroscience}, 12:126, 2018.

\bibitem{carvalho2009differential}
Tiago~P Carvalho and Dean~V Buonomano.
\newblock Differential effects of excitatory and inhibitory plasticity on
  synaptically driven neuronal input-output functions.
\newblock {\em Neuron}, 61(5):774--785, 2009.

\bibitem{hofer2011differential}
Sonja~B Hofer, Ho~Ko, Bruno Pichler, Joshua Vogelstein, Hana Ros, Hongkui Zeng,
  Ed~Lein, Nicholas~A Lesica, and Thomas~D Mrsic-Flogel.
\newblock Differential connectivity and response dynamics of excitatory and
  inhibitory neurons in visual cortex.
\newblock {\em Nature neuroscience}, 14(8):1045--1052, 2011.

\bibitem{sjostrom2008dendritic}
P~Jesper Sjostrom, Ede~A Rancz, Arnd Roth, and Michael Hausser.
\newblock Dendritic excitability and synaptic plasticity.
\newblock {\em Physiological reviews}, 88(2):769--840, 2008.

\bibitem{feldman2012spike}
Daniel~E Feldman.
\newblock The spike-timing dependence of plasticity.
\newblock {\em Neuron}, 75(4):556--571, 2012.

\bibitem{korte2016cellular}
Martin Korte and Dietmar Schmitz.
\newblock Cellular and system biology of memory: timing, molecules, and beyond.
\newblock {\em Physiological reviews}, 96(2):647--693, 2016.

\bibitem{pool2011spike}
R~Rossi Pool and Germ{\'a}n Mato.
\newblock Spike-timing-dependent plasticity and reliability optimization: the
  role of neuron dynamics.
\newblock {\em Neural computation}, 23(7):1768--1789, 2011.

\bibitem{legenstein2007edge}
Robert Legenstein and Wolfgang Maass.
\newblock Edge of chaos and prediction of computational performance for neural
  circuit models.
\newblock {\em Neural networks}, 20(3):323--334, 2007.

\bibitem{frazier2018tutorial}
Peter~I Frazier.
\newblock A tutorial on bayesian optimization.
\newblock {\em arXiv preprint arXiv:1807.02811}, 2018.

\bibitem{eriksson2021high}
David Eriksson and Martin Jankowiak.
\newblock High-dimensional bayesian optimization with sparse axis-aligned
  subspaces.
\newblock In {\em Uncertainty in Artificial Intelligence}, pages 493--503.
  PMLR, 2021.

\bibitem{feydy2019interpolating}
Jean Feydy, Thibault S{\'e}journ{\'e}, Fran{\c{c}}ois-Xavier Vialard, Shun-ichi
  Amari, Alain Trouv{\'e}, and Gabriel Peyr{\'e}.
\newblock Interpolating between optimal transport and mmd using sinkhorn
  divergences.
\newblock In {\em The 22nd International Conference on Artificial Intelligence
  and Statistics}, pages 2681--2690. PMLR, 2019.

\bibitem{wong2020tinyspeech}
Alexander Wong, Mahmoud Famouri, Maya Pavlova, and Siddharth Surana.
\newblock Tinyspeech: Attention condensers for deep speech recognition neural
  networks on edge devices.
\newblock {\em arXiv preprint arXiv:2008.04245}, 2020.

\bibitem{chakraborty2021fully}
Biswadeep Chakraborty, Xueyuan She, and Saibal Mukhopadhyay.
\newblock A fully spiking hybrid neural network for energy-efficient object
  detection.
\newblock {\em arXiv preprint arXiv:2104.10719}, 2021.

\bibitem{markram1997regulation}
Henry Markram, Joachim L{\"u}bke, Michael Frotscher, and Bert Sakmann.
\newblock Regulation of synaptic efficacy by coincidence of postsynaptic aps
  and epsps.
\newblock {\em Science}, 275(5297):213--215, 1997.

\bibitem{wang2019space}
Qinyi Wang, Yexin Zhang, Junsong Yuan, and Yilong Lu.
\newblock Space-time event clouds for gesture recognition: From rgb cameras to
  event cameras.
\newblock In {\em 2019 IEEE Winter Conference on Applications of Computer
  Vision (WACV)}, pages 1826--1835. IEEE, 2019.

\bibitem{bi2020graph}
Yin Bi, Aaron Chadha, Alhabib Abbas, Eirina Bourtsoulatze, and Yiannis
  Andreopoulos.
\newblock Graph-based spatio-temporal feature learning for neuromorphic vision
  sensing.
\newblock {\em IEEE Transactions on Image Processing}, 29:9084--9098, 2020.

\bibitem{carreira2017quo}
Joao Carreira and Andrew Zisserman.
\newblock Quo vadis, action recognition? a new model and the kinetics dataset.
\newblock In {\em proceedings of the IEEE Conference on Computer Vision and
  Pattern Recognition}, pages 6299--6308, 2017.

\bibitem{lee2021low}
Hyunhoon Lee, Young-Seok Kim, Mijung Kim, and Youngjoo Lee.
\newblock Low-cost network scheduling of 3d-cnn processing for embedded action
  recognition.
\newblock {\em IEEE Access}, 9:83901--83912, 2021.

\bibitem{wang2021tdn}
Limin Wang, Zhan Tong, Bin Ji, and Gangshan Wu.
\newblock Tdn: Temporal difference networks for efficient action recognition.
\newblock In {\em Proceedings of the IEEE/CVF Conference on Computer Vision and
  Pattern Recognition}, pages 1895--1904, 2021.

\bibitem{zheng2020going}
Hanle Zheng, Yujie Wu, Lei Deng, Yifan Hu, and Guoqi Li.
\newblock Going deeper with directly-trained larger spiking neural networks.
\newblock {\em arXiv preprint arXiv:2011.05280}, 2020.

\bibitem{shen2021backpropagation}
Guobin Shen, Dongcheng Zhao, and Yi~Zeng.
\newblock Backpropagation with biologically plausible spatio-temporal
  adjustment for training deep spiking neural networks.
\newblock {\em arXiv preprint arXiv:2110.08858}, 2021.

\bibitem{liu2021event}
Qianhui Liu, Dong Xing, Huajin Tang, De~Ma, and Gang Pan.
\newblock Event-based action recognition using motion information and spiking
  neural networks.
\newblock In {\em Proceedings of the Thirtieth International Joint Conference
  on Artificial Intelligence, IJCAI-21, Z.-H. Zhou, Ed. International Joint
  Conferences on Artificial Intelligence Organization}, volume~8, pages
  1743--1749, 2021.

\bibitem{patravali2021unsupervised}
Jay Patravali, Gaurav Mittal, Ye~Yu, Fuxin Li, and Mei Chen.
\newblock Unsupervised few-shot action recognition via action-appearance
  aligned meta-adaptation.
\newblock In {\em Proceedings of the IEEE/CVF International Conference on
  Computer Vision}, pages 8484--8494, 2021.

\bibitem{soomro2017unsupervised}
Khurram Soomro and Mubarak Shah.
\newblock Unsupervised action discovery and localization in videos.
\newblock In {\em Proceedings of the IEEE International Conference on Computer
  Vision}, pages 696--705, 2017.

\bibitem{meng2011modeling}
Yan Meng, Yaochu Jin, and Jun Yin.
\newblock Modeling activity-dependent plasticity in bcm spiking neural networks
  with application to human behavior recognition.
\newblock {\em IEEE transactions on neural networks}, 22(12):1952--1966, 2011.

\bibitem{de1987feedback}
ER~De~Kloet and JMHM Reul.
\newblock Feedback action and tonic influence of corticosteroids on brain
  function: a concept arising from the heterogeneity of brain receptor systems.
\newblock {\em Psychoneuroendocrinology}, 12(2):83--105, 1987.

\bibitem{petitpre2018neuronal}
Charles Petitpr{\'e}, Haohao Wu, Anil Sharma, Anna Tokarska, Paula Fontanet,
  Yiqiao Wang, Fran{\c{c}}oise Helmbacher, Kevin Yackle, Gilad Silberberg,
  Saida Hadjab, et~al.
\newblock Neuronal heterogeneity and stereotyped connectivity in the auditory
  afferent system.
\newblock {\em Nature communications}, 9(1):1--13, 2018.

\bibitem{shamir2006implications}
Maoz Shamir and Haim Sompolinsky.
\newblock Implications of neuronal diversity on population coding.
\newblock {\em Neural computation}, 18(8):1951--1986, 2006.

\bibitem{tu2021dimensionality}
Chengyi Tu, Paolo D'Odorico, and Samir Suweis.
\newblock Dimensionality reduction of complex dynamical systems.
\newblock {\em Iscience}, 24(1):101912, 2021.

\bibitem{gao2016universal}
Jianxi Gao, Baruch Barzel, and Albert-L{\'a}szl{\'o} Barab{\'a}si.
\newblock Universal resilience patterns in complex networks.
\newblock {\em Nature}, 530(7590):307--312, 2016.

\bibitem{kubota2021unifying}
Tomoyuki Kubota, Hirokazu Takahashi, and Kohei Nakajima.
\newblock Unifying framework for information processing in stochastically
  driven dynamical systems.
\newblock {\em Physical Review Research}, 3(4):043135, 2021.

\bibitem{jayasumana2015kernel}
Sadeep Jayasumana, Richard Hartley, Mathieu Salzmann, Hongdong Li, and Mehrtash
  Harandi.
\newblock Kernel methods on riemannian manifolds with gaussian rbf kernels.
\newblock {\em IEEE transactions on pattern analysis and machine intelligence},
  37(12):2464--2477, 2015.

\bibitem{kolouri2016sliced}
Soheil Kolouri, Yang Zou, and Gustavo~K Rohde.
\newblock Sliced wasserstein kernels for probability distributions.
\newblock In {\em Proceedings of the IEEE Conference on Computer Vision and
  Pattern Recognition}, pages 5258--5267, 2016.

\bibitem{kolouri2015radon}
Soheil Kolouri, Se~Rim Park, and Gustavo~K Rohde.
\newblock The radon cumulative distribution transform and its application to
  image classification.
\newblock {\em IEEE transactions on image processing}, 25(2):920--934, 2015.

\bibitem{mejias2014differential}
Jorge~F Mejias and Andr{\'e} Longtin.
\newblock Differential effects of excitatory and inhibitory heterogeneity on
  the gain and asynchronous state of sparse cortical networks.
\newblock {\em Frontiers in computational neuroscience}, 8:107, 2014.

\end{thebibliography}

\end{document}